%% file: _main.tex
\input{_constants}
\arxiv

\pdfoutput=1
\documentclass[10pt,twocolumn,letterpaper]{article}
\input{cvpr_header}

\unless\ifarxiv \myexternaldocument{_supplementary} \fi

\begin{document}
\title{\paperTitle}
\author{\authorBlock}

\maketitle

\input{00_abstract}
\input{01_intro}

\input{02_related}

\input{03_method}
\input{04_experiments}
\input{05_conclusion}

{\small
\bibliographystyle{ieeenat_fullname}
\bibliography{06_references}
}

\ifarxiv
\clearpage
\appendix
\maketitlesupplementary
\input{07_appendix}

\fi

\end{document}

%% file: _constants.tex
\def\paperID{}
\def\confName{CVPR}
\def\confYear{2026}

\def\paperTitle{Query2Uncertainty: Robust Uncertainty Quantification and Calibration for \\ 3D Object Detection under Distribution Shift}

\def\authorBlock{
    Till Beemelmanns$^{1}$ \qquad
    Alexey Nekrasov$^{2}$ \qquad
    Stefan Vilceanu$^{1}$ \qquad
    Jonas Steinhaus$^{1}$ \qquad \\
    Timo Woopen$^{1}$ \qquad
    Bastian Leibe$^{2}$ \qquad
    Lutz Eckstein$^{1}$ \\
    \normalsize $^{1}$Institute for Automotive Engineering, RWTH Aachen \quad $^{2}$Computer Vision Institute, RWTH Aachen \\
}

\newif\ifreview 
\newif\ifarxiv \newcommand{\arxiv}{\arxivtrue}
\newif\ifcamera 
\newif\ifrebuttal 

%% file: cvpr_header.tex
\ifreview \usepackage[review]{cvpr} \fi
\ifarxiv \usepackage[pagenumbers]{cvpr} \fi
\ifrebuttal \usepackage[rebuttal]{cvpr} \fi
\ifcamera \usepackage{cvpr} \fi

\input{_macros}  %

\usepackage{xr-hyper}

\makeatletter
\newcommand*{\addFileDependency}[1]{
  \typeout{(#1)}
  \@addtofilelist{#1}
  \IfFileExists{#1}{}{\typeout{No file #1.}}
}

\makeatother
\newcommand*{\myexternaldocument}[1]{
    \externaldocument{#1}
    \addFileDependency{#1.tex}
    \addFileDependency{#1.aux}
}

\definecolor{cvprblue}{rgb}{0.21,0.49,0.74}
\usepackage[pagebackref,breaklinks,colorlinks,allcolors=cvprblue]{hyperref}
\usepackage[capitalize]{cleveref}
\crefname{section}{Sec.}{Secs.}
\crefname{table}{Table}{Tables}
\crefname{figure}{Fig.}{Figs.}

\ifarxiv \crefname{appendix}{App.}{Apps.}
\else \crefname{appendix}{Suppl.}{Suppls.} \fi

%% file: _macros.tex
\usepackage{graphicx}
\usepackage{amsmath}
\usepackage{amssymb}
\usepackage{booktabs}
\usepackage{times}
\usepackage{microtype}
\usepackage{epsfig}
\usepackage{caption}
\usepackage{float}
\usepackage{placeins}
\usepackage{color, colortbl}
\usepackage{stfloats}
\usepackage{enumitem}
\usepackage{tabularx}
\usepackage{xstring}
\usepackage{multirow}
\usepackage{xspace}
\usepackage{url}
\usepackage{subcaption}
\usepackage{xcolor, xfp}
\usepackage[hang,flushmargin]{footmisc}
\usepackage{lipsum}
\usepackage{tabularx}
\usepackage{makecell}
\usepackage[export]{adjustbox}

\usepackage{tikz}
\usepackage{pgfplots}
\usepackage{pgfplotstable}
\usepgfplotslibrary{groupplots}
\usepgfplotslibrary{statistics}
\usetikzlibrary{shapes.geometric}
\usepackage{listings}

\definecolor{codegreen}{rgb}{0,0.6,0}
\definecolor{codegray}{rgb}{0.5,0.5,0.5}
\definecolor{codepurple}{rgb}{0.58,0,0.82}
\definecolor{backcolour}{rgb}{0.95,0.95,0.92}

\lstdefinestyle{mystyle}{
    backgroundcolor=\color{backcolour},
    commentstyle=\color{codegreen},
    keywordstyle=\color{magenta},
    numberstyle=\tiny\color{codegray},
    stringstyle=\color{codepurple},
    basicstyle=\ttfamily\footnotesize,
    breakatwhitespace=false,
    breaklines=true,
    captionpos=b,
    keepspaces=true,
    numbers=left,
    numbersep=5pt,
    showspaces=false,
    showstringspaces=false,
    showtabs=false,
    tabsize=1
}
\usepackage[most]{tcolorbox}
\newtcbox{\globalbox}{
  on line,
  size=fbox,
  tcbox raise=-0.4mm,
  colupper=black!95!white,
  enhanced,
  fontupper=\scriptsize,
  boxsep=1.5pt,
  left=0pt,
  right=0pt,
  top=0pt,
  bottom=0pt,
  boxrule=0.25mm,
  colframe=black!60!white,
  colback=magenta!20!white}
\newcommand{\gbox}{\globalbox{Glb.}}

\newtcbox{\classbox}{
  on line,
  size=fbox,
  tcbox raise=-0.4mm,
  colupper=black!95!white,
  enhanced,
  fontupper=\scriptsize,
  boxsep=1.5pt,
  left=0pt,
  right=0pt,
  top=0pt,
  bottom=0pt,
  boxrule=0.25mm,
  colframe=black!60!white,
  colback=cyan!20!white}
\newcommand{\cbox}{\classbox{Cls.}}

\newcommand{\deltatuck}{-0.00em}

\newcommand{\da}[2][2]{%
  \begingroup
  \edef\s{\fpeval{trunc(sign(#2))}}%
  \kern\deltatuck
  \ifnum\s=0\relax
    \rlap{\textsubscript{\smash{\textcolor{gray!70!black}{\scriptsize o}}}}%
  \else\ifnum\s>0\relax
    \rlap{\textsubscript{\smash{\textcolor{green!60!black}{\scriptsize +}}}}%
  \else
    \rlap{\textsubscript{\smash{\textcolor{red!65!black}{\scriptsize -}}}}%
  \fi\fi
  \endgroup
}

\renewcommand{\paragraph}[1]{\vspace{.5em}\noindent\textbf{#1.}}

\usepackage{soul}
\setuldepth{foobar}

\renewcommand{\bf}[1]{\textbf{#1}}

\newcommand{\un}[1]{\underline{#1}}

\newcommand{\supp}{supplemental material\xspace}
\ifarxiv \renewcommand{\supp}{appendix\xspace} \fi

\newcommand{\R}[1]{{%
    \textbf{%
        \ifstrequal{#1}{1}{\textcolor{red}{R#1}}{%
        \ifstrequal{#1}{2}{\textcolor{blue}{R#1}}{%
        \ifstrequal{#1}{3}{\textcolor{magenta}{R#1}}{%
        \ifstrequal{#1}{4}{\textcolor{teal}{R#1}}{%
                           \textcolor{cyan}{R#1}%
        }}}}%
    }%
}}

\newcommand{\cb}{\cellcolor{blue!10}}

\definecolor{m_green_border}{HTML}{82B366}
\definecolor{m_darkgreen}{HTML}{d5e9d5}
\definecolor{m_green}{HTML}{d5e9d5}
\definecolor{m_orange}{HTML}{fff2cd}
\definecolor{m_orange_border}{RGB}{150,114,164}
\definecolor{m_red}{HTML}{fccecd}
\definecolor{m_violet}{HTML}{e2d6e8}
\definecolor{m_blue}{HTML}{d8e9fc}
\definecolor{m_queries}{RGB}{213,232,212}
\definecolor{m_queries_border}{RGB}{130,179,102}
\definecolor{m_red}{RGB}{248,206,204}
\definecolor{m_red_border}{RGB}{204,29,12}
\definecolor{m_z_dens}{RGB}{255,223,209}
\definecolor{m_z_dens_border}{RGB}{250,139,139}
\definecolor{m_blue_border}{RGB}{107,141,190}
\definecolor{m_yellow}{RGB}{255,242,205}
\definecolor{m_yellow_border}{RGB}{213,182,82}
\definecolor{m_gray}{RGB}{245,245,245}
\definecolor{m_gray_border}{RGB}{102,102,102}
\newcommand{\colorsquare}[1]{\tikz{\path[draw=#1_border,fill=#1, thick, rounded corners=0.6pt] (0,0) rectangle (6pt,6pt);}}

\newcommand{\PAR}[1]{\vskip4pt \noindent {\textbf{#1}~}}
\newcommand{\PARbegin}[1]{\noindent {\textbf{#1}~}}

%% file: 00_abstract.tex
\begin{abstract}
Reliable uncertainty estimation for 3D object detection is critical for deploying safe autonomous systems, yet modern detectors remain poorly calibrated, especially under distribution shifts.
Although post-hoc calibration methods address this issue and provide improved calibration for in-distribution tests, they fail to adapt in distribution-shifted scenarios.
In this work, we address this issue and introduce a density-aware calibration method that couples post-hoc calibrators with the feature density of latent object queries from DETR-style 3D object detectors.
These queries form a compact, location and class-aware feature, ideal for density estimation, allowing our approach to adjust model confidences in distribution-shift scenarios.
By fitting a density estimator on these query features, our approach jointly recalibrates both classification and bounding box regression uncertainties.
On both a multi-view camera and LiDAR-based detector, our approach consistently outperforms standard post-hoc methods in both in-distribution and distribution-shifted scenarios.
Code: \url{https://tillbeemelmanns.github.io/query2uncertainty/}.
\end{abstract}
\vspace{-1em}

%% file: 01_intro.tex
\section{Introduction}
Autonomous vehicles and robots are nowadays deployed in real-world scenarios and provide numerous services and benefits to society.
These systems must operate under constant uncertainty, arising from sensor imperfections, environmental variability, occlusion, or unseen data.
Perceiving an accurate environment representation from a robot's sensor system, along with precise uncertainty estimates, is crucial for autonomous vehicles and robots to navigate and interact with other agents.
Current methods for autonomous driving use multi-view camera images and LiDAR point clouds to detect nearby objects~\cite{liu2022petr,liu2022petrv2,centerformer,yan2023cross}.
These objects are usually detected as 3D bounding boxes that are associated with a semantic class, the object's center position, its size and orientation, and serve as input for object tracking and collision avoidance~\cite{Weng2020_AB3DMOT}. 
Each of the box parameters can be associated with an uncertainty measure, improving probabilistic trackers or V2X setups to better fuse information from multiple sources or noisy measurements \cite{coalign,kld}.
However, DNN-based detectors are known to produce overconfident predictions that might affect downstream performance and pose serious safety risks in real-world operation~\cite{ts,kuzucu2024calibration}.
\input{figs/teaser.tex}
In this work, we focus on obtaining reliable uncertainty estimates for modern 3D object detectors, especially in unseen scenarios.
Recent approaches focus on increasing detector accuracy through architectural model improvements in fusing multiple cameras, through temporal fusion, or by combining multiple modalities~\cite{liu2022petr,yan2023cross,liu2022petrv2,li2022bevformer}; however, retrieving and evaluating uncertainty estimates of predicted location and class probabilities is often not considered.
This poses a significant problem for deploying 3D detectors in safety‑critical applications: miscalibrated confidences and variances sent to other agents, or propagated to the tracking and planning stack, might lead to brittle behavior and elevated risk.
Current methods for calibrating class scores in 2D detectors use training losses to steer the model to predict reliable confidences~\cite{caldetr,tcd} or use post‑hoc calibration~\cite{ts,ps,ir,kuzucu2024calibration}.
Post-hoc methods for detection have been shown to outperform training-time calibration~\cite{kuzucu2024calibration}, but in-distribution calibrated post-hoc methods, in-turn, fail in domain-shift and out-of-distribution (OOD) scenarios~\cite{tomani2021,tomani2023}.
So far, no approach has been proposed to solve the problem of obtaining robust calibrated uncertainties for probabilistic 3D bounding box detections in in-distribution and distribution-shift scenarios.
A recent line of work proposes to calibrate classifier confidences based on feature density~\cite{tomani2021,tomani2023} for image classification.
Query-based 3D object detectors expose latent object hypotheses, which drive detection of class, location, size, and orientation.
We use these latent object hypotheses to fit a density estimator and use the resulting density to make post-hoc calibrators feature density-aware.
Our approach yields a density-aware calibration module that is lightweight and adaptive to unseen data.

Although there exists a wide range of metrics that quantify uncertainty~\cite{difeng2022,sicking2019,kuleshov2018accurate,es,dece,ts}, there is no unified benchmark that comprehensively evaluates the quality of uncertainty in 3D object detection.
We therefore propose a standardized evaluation framework for uncertainty estimates in 3D object detection, built on top of the widely used nuScenes detection benchmark~\cite{caesar2020nuscenes}.
We evaluate our proposed density-aware uncertainty calibration method with a camera and LiDAR-based model and demonstrate superior performance across in-distribution and distribution-shifted scenarios for regression and classification uncertainties.

Our main contributions are as follows:
    1) We establish a comprehensive uncertainty quantification benchmark for 3D object detection. 
    2) We propose a density-aware post-hoc calibration method that couples feature density with classic calibrators to refine both classification and regression uncertainty estimates of query-based 3D detectors.
    3) We empirically demonstrate on both a camera and LiDAR-based model, that our proposed approach outperforms baseline methods across in-distribution and distribution-shift settings.

%% file: figs/teaser.tex
\begin{figure}[tp]
    \centering
    \includegraphics[width=0.84\linewidth]{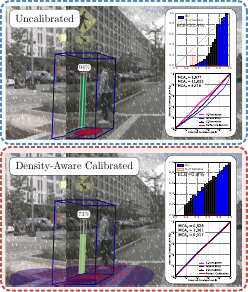}
    \caption{
    Our density-aware calibration method improves reliability under distribution-shift.
    An uncalibrated 3D detector exhibits miscalibrated uncertainty estimates when facing distribution shifts, such as a heavy rain or snow.
    Our method uses feature density to perform density-aware calibration of classification confidences and regression values, improving the model's trustworthiness.
    }
    \label{fig:teaser}
\end{figure}

%% file: 02_related.tex
\section{Related Work}\label{sec:related}

\PARbegin{Uncertainty Quantification in 3D Object Detection.}
Modern 3D object detection for autonomous driving has rapidly adopted the query-based paradigm from 2D models like DETR \cite{Carion2020}.
Multi-view camera models~\cite{Wang2022,liu2022petr,li2022bevformer,liu2022petrv2,doll2022,wang2023exploring} and LiDAR-based detectors~\cite{centerformer,erabati2023li3detr,liu2024seed} rely on object queries for object classification and bounding box regression.
Multiple works have explored methods to estimate the predictive uncertainty, such as Monte-Carlo Dropout (MCD)~\cite{mcd,gal2017,kendall2017uncertainties,postels2019}, ensemble methods~\cite{de}, or parametric models directly modeling the variance~\cite{nix1994,heskes1996}.
In the field of autonomous driving, Monte Carlo Dropout~\cite{harakeh2020,difeng2019mcd}, MIMO~\cite{pitropov2022}, or the use of additional heads for directly regressing box parameter variances, trained with a KL-divergence or energy-based losses \cite{gaussianyolo,he2018,es,kld,coalign,trust} have been explored extensively.
However, sampling-based methods are often inefficient, and parametric methods typically focus on either classification or regression, but not an unified calibration.
We propose a method that jointly quantifies and calibrates both classification and bounding box regression uncertainty for query-based 3D detectors.

\PAR{Post-hoc Calibration Methods.}
Estimating predictive uncertainty alone is insufficient, as modern models' confidence scores do not accurately reflect the likelihood of a prediction.
The most common solution is post-hoc calibration, which adjusts the model's outputs after training without costly re-training.
Nonparametric methods, which do not assume an underlying distribution, use Histogram Binning~\cite{zadrozny2001}, Isotonic Regression~\cite{ir}, or Bayesian Binning into Quantiles~\cite{PakdamanNaeini2015}.
In contrast, parametric post-hoc methods assume an underlying Gaussian distribution and adjust logits before applying a sigmoid or softmax, as popularized by Platt Scaling (PS)~\cite{ps} and Temperature Scaling (TS)~\cite{ts}.
Beyond these classical primitives, recent works add structural priors to the calibration.
For example, depth-aware strategies for 3D point cloud segmentation couple confidence with depth~\cite{calib3d}.
While most approaches target classification, TS has also been applied to regression variances for 3D box parameters~\cite{trust}.
While these methods can effectively correct miscalibration on in-distribution data, they are fundamentally ill-equipped to handle distribution shifts, as they blindly apply the same correction to all inputs, assuming the test data follows the same data distribution.

\PAR{Calibration under Distribution Shift.}
The primary weakness of standard post-hoc methods is their failure under distribution shift; a calibrator tuned on in-distribution data will collapse when faced with out-of-distribution (OOD) samples \cite{tomani2021,tomani2023}.
This motivates the need for calibration methods that are robust to distribution shifts.
One line of work attempts to solve this during training~\cite{tcd,pathiraja2023multiclass,caldetr}, but these methods often focus on classification calibration for 2D detection, and a recent analysis shows that carefully tuned post-hoc baselines can outperform them \cite{kuzucu2024calibration}.
A more promising direction, which we build upon, is calibration based on the feature density, commonly used in the OOD setting~\cite{du2022vos,liang2022gmmseg,lee2018simple,heidrich2025occuq}.
This paradigm, applied successfully in image classification \cite{tomani2021,tomani2023}, dynamically adjusts the calibration based on estimating the feature density the prediction is sampled from.
By modeling feature density, these methods can identify OOD inputs and correct their confidences, achieving much better calibration, especially under 
distribution shift.
However, extending this density-aware principle to probabilistic 3D object detection---to jointly recalibrate both classification and bounding box regression uncertainties---has not been explored.
We propose a method that models the feature density of object queries to make post-hoc calibration robust under adverse scenarios.

%% file: 03_method.tex
\section{Method}\label{sec:method}
To introduce robust regression and classification calibration for DETR-style 3D object detectors, we propose to couple feature density-based uncertainty quantification with post-hoc calibration.
We establish an uncertainty evaluation framework tailored for nuScenes~\cite{caesar2020nuscenes}, introducing metrics that assess both classification and regression uncertainties.
Our approach outperforms established baselines for in-distribution and distribution-shifted scenarios.
\input{figs/method_overview}

\subsection{Method Overview}
From 3D sensor data (either multi-view camera or point cloud), our method first extracts features using standard backbones and applies positional encodings to retain spatial information.
Following DETR-style 3D detector pipelines, we interpret these features as tokens (Figure~\ref{fig:method_overview} \colorsquare{m_yellow}) and feed them into our probabilistic decoder-only transformer, where they interact with a set of learnable object queries $z_0$~(\colorsquare{m_queries}).
The decoder outputs refined queries $z$~(\colorsquare{m_queries}), which are then used to predict class scores, mean box parameters, and associated variances using separate detection heads~(\colorsquare{m_blue}), obtaining boxes in the form of $\mathbf{b}=(\hat{x},\hat{y},\hat{z},\hat{\ell},\hat{w},\hat{h},\hat{\theta},\hat{v}_x,\hat{v}_y,\sigma_x^2,\sigma_y^2,\sigma_z^2,\sigma_{\ell}^2,\sigma_w^2,\sigma_h^2,\kappa_\theta)$.
Simultaneously, we use an epistemic density estimation module that models how well the final object queries align with True Positive queries from the training set.
At test time, we compute the density of each object query feature~(\colorsquare{m_z_dens}) and use this information to dynamically calibrate both the classification and regression uncertainties in our density-aware post-hoc calibration module~(\colorsquare{m_red}).
For our experiments, we build upon the DETR-style 3D detector PETR~\cite{liu2022petr}, which processes multi-view camera images.
To further validate our approach, we use a LiDAR-based detector and a SECOND point cloud encoder~\cite{yan2018second} instead of the camera-based encoder while keeping the rest of the architecture unchanged.

\subsection{Uncertainty Evaluation Framework}
To evaluate our approach, we extend the nuScenes detection benchmark~\cite{caesar2020nuscenes} with uncertainty evaluation for both classification and regression and provide an interface to use this benchmark (See \supp for more detail).

\PAR{Classification Uncertainty.}
Evaluating a classifier's calibration aims to assess the alignment between expected accuracy and actual confidence.
It is usually measured using the Expected Calibration Error (ECE)~\cite{ts}, which measures the difference between expected accuracy and actual confidence over a set of discrete bins.
For object detection, the \emph{Detection Expected Calibration Error} (D-ECE)~\cite{dece} adapts the ECE to measure the discrepancy between precision and predicted confidence.
We discretize the predicted confidences into $B$ bins and denote by $\mathcal{D}_b$ the detections whose confidence falls into bin $b$.
The average confidence in bin $b$ is $\overline{s}_b = \frac{1}{|\mathcal{D}_b|} \sum_{j \in \mathcal{D}_b} s_j$.
True Positives (TPs) are identified via the nuScenes Euclidean center distance matching rule \cite{caesar2020nuscenes}, forming the subset $\mathcal{T}_b \subseteq \mathcal{D}_b$.
The bin precision therefore is $\pi_b = \frac{|\mathcal{T}_b|}{|\mathcal{D}_b|}$ and the bin weight reads $w_b = |\mathcal{D}_b| / n$ with $n = \sum_{b=1}^B |\mathcal{D}_b|$.
Empty bins receive $w_b = 0$ and do not contribute. D-ECE then aggregates the discrepancy between confidence and precision,
\begin{equation}%
    \text{D-ECE} = \sum_{b=1}^{B} w_b \left| \overline{s}_b - \pi_b \right|.
\end{equation}
In contrast to related work~\cite{caldetr,kuzucu2024calibration}, we follow the nuScenes definition of TPs based on the Euclidean center distance between predicted and ground-truth centroids, instead of the IoU.
To better reflect the interplay between confidence and localization quality, we adopt the \emph{Location-Aware Expected Calibration Error} (LaECE)~\cite{saod}.
Using the same binning as for D-ECE, we compute the average confidence $\overline{s}_b$ and a location quality term.
Matched detections in bin $b$ contribute with $lq_{j} = 1 - \frac{\min(d_j, \tau)}{\tau}$, where $d_j$ is the Euclidean center distance between the predicted and ground-truth boxes and $\tau = 2\mathrm{m}$ denotes the nuScenes matching threshold.
False Positives (FPs) receive $lq_{j} = 0$, such that missed localization performance diminishes the bin's score.
By averaging the localization quality within each bin and directly plugging it into the LaECE definition we obtain
\begin{equation}%
    \text{LaECE} = \sum_{b=1}^{B} w_b \left| \overline{s}_b - \frac{1}{|\mathcal{D}_b|} \sum_{j \in \mathcal{T}_b} lq_{j} \right|,
\end{equation}
where $\mathcal{D}_b$ denotes all detections (TPs and FPs) in bin $b$, $\mathcal{T}_b$ the matched detections, and $w_b = |\mathcal{D}_b| / n$ weighs each bin by its share of detections.
This metric penalizes confidence estimates that do not align with the achieved spatial accuracy and rewards models producing confident well-localized predictions.
We use $B=25$ bins for both LaECE and D-ECE.
The benchmark also implements other metrics such as the LaACE and Uncertainty Realism Criterion~\cite{sicking2019}, as described in the \supp.

\PAR{Regression Uncertainty.}
We assess Gaussian predictive uncertainties using the \emph{Miscalibration Area} (MCA)~\cite{kuleshov2018accurate}, the area between nominal and empirical coverage for \emph{centered} prediction intervals, as a pure measure of calibration quality.
Let $\{(\mu_i,\sigma_i^2)\}_{i=1}^{N}$ denote the scalar Gaussian posterior means and variances for TPs with ground truth $y_{i,\text{gt}}$. We define standardized residuals
\begin{equation}
    r_i=\frac{y_{i,\text{gt}}-\mu_i}{\sigma_i}.
\end{equation}
For a nominal coverage level $p\in[0,1]$, the symmetric standard-normal threshold reads
\begin{equation}
    s(p)=\Phi^{-1}\!\big(0.5+\tfrac{p}{2}\big),
\end{equation}
where $\Phi$ is the CDF of the standard normal and $\Phi^{-1}$ its quantile (probit) function. The empirical coverage at level $p$ is
\begin{equation}
    \widehat{\pi}(p)=\frac{1}{N}\sum_{i=1}^{N}\mathbb{I}\{|r_i|\le s(p)\},
\end{equation}
where $\mathbb{I}\{\cdot\}$ denotes the indicator function. The miscalibration area then reads
\begin{equation}
    \mathrm{MCA}=\int_{0}^{1}\big|\widehat{\pi}(p)-p\big|\,\mathrm{d}p.
\end{equation}
We approximate the integral via a trapezoidal rule, using $B=100$ bins to discretize the probability space $[0,1]$.
We report MCA separately for centroid ($\mathrm{MCA}_{xyz}$) and size ($\mathrm{MCA}_{lwh}$), where we take the average over the three dimensions.
For the angular component, we wrap predictions and targets to $[-\pi,\pi)$. %

\PAR{Accounting for Class Imbalance.}
The nuScenes detection dataset exhibits severe class imbalance~\cite{caesar2020nuscenes}, so naïve metrics are dominated by most frequent categories. %
We therefore report each metric as a \emph{class-wise average}, to ensure that rare classes contribute equally.

\subsection{Probabilistic Transformer Detector}
We parameterize each 3D box with uncertainty as $\mathbf{b} = (\hat{x},\hat{y},\hat{z},\hat{\ell},\hat{w},\hat{h},\hat{\theta},\hat{v}_x,\hat{v}_y,\sigma_x^2,\sigma_y^2,\sigma_z^2,\sigma_{\ell}^2,\sigma_w^2,\sigma_h^2,\kappa_\theta)$, coupling the 3D center, size and yaw with uncertainty.

\PAR{Regression Uncertainty Head.}
We supervise a separate uncertainty head via heteroscedastic losses derived from the KL divergence between the predicted distribution and a ground-truth Dirac distribution~\cite{coalign,kld}.
We condition the MLP head to minimize the KL divergence between the estimated Gaussian $\mathcal{N}(x;\hat{x},\sigma_x^2)$ and the ground-truth delta distribution $P(x) = \delta(x - x_0)$, which yields $\mathcal{L}_x = (\hat{x} - x_0)^2 /(2\sigma_x^2) + \log\sigma_x^2/2$.
For the box center coordinates $(x,y,z)$ we assume independent Gaussians with log-variance $u_i$, leading to the loss
\begin{equation}
    \mathcal{L}_{xyz} = \frac{1}{2}\sum_{i\in\{x,y,z\}}\left((\hat{x}_i - x_{0,i})^2 e^{-u_i} + u_i\right),
\end{equation}
where $\hat{x}_i$ and $x_{0,i}$ denote prediction and ground-truth, respectively.
The size parameters $(\ell,w,h)$ are predicted in log-space; we exponentiate before measuring the residual, so the analogous loss reads
\begin{equation}
    \mathcal{L}_{\ell wh} = \frac{1}{2}\sum_{i\in\{\ell,w,h\}}\left((e^{\hat{x}_i} - e^{x_{0,i}})^2 e^{-u_i} + u_i\right).
\end{equation}
To model the yaw angle, we consider a von-Mises distribution with mean $\hat{\theta}$ and concentration $\kappa = e^{-u_\theta} = 1/\sigma_\theta^2$; minimizing the KL divergence between $\mathrm{vM}(\theta;\hat{\theta},1/\sigma_\theta^2)$ and $P(\theta) = \delta(\theta - \theta_0)$ yields
\begin{equation}
    \mathcal{L}_{\theta} = \log I_0(e^{-u_\theta}) + e^{-u_\theta}\bigl[1 - \cos(\hat{\theta}-\theta_0)\bigr],
\end{equation}
where $I_0$ is the modified Bessel function. For numerical stability we add the term $\lambda_V\,\mathrm{ELU}(u_\theta - s_0)$~\cite{coalign}, where $\lambda_V$ controls the penalty strength and $s_0$ sets the ELU offset to keep gradients well-behaved for small concentrations.

\PAR{Query-Density Estimator Training.}
We propose to estimate the feature density of the query vectors in the latent space to capture how well a query aligns with the training distribution.
To train this model, we conduct a forward pass on the training split and cache the latent query vectors from the last transformer decoder layer which correspond to TP matches.
For each semantic class $c$ we aggregate the matched queries, count the samples $N_c$, and form the empirical prior $\hat{\omega}_c = N_c / \sum_{k} N_k$.
We then fit a RealNVP normalizing flow~\cite{dinh2017density} to each class-conditioned feature set.
Every flow stacks a diagonal Gaussian base distribution with 32 affine coupling blocks interleaved with swap permutations; the scale and shift in each block are predicted by two-layer MLPs with 32 hidden units and zero-initialized outputs to stabilize training.
We optimize the flows for 60 epochs using Adam with a linear learning-rate decay.
The query feature is a vector of dimension $D=256$.
The objective is the forward $\mathrm{KL}(p\|q)$ between the empirical feature distribution and the flow estimator.
We minimize the negative log-likelihood averaged over a batch of $M$ query samples $\{z_i\}_{i=1}^{M}$:
\begin{equation}\label{eq:10}
    \mathcal{L}_{\text{flow}} = - \frac{1}{M} \sum_{i=1}^{M} \log q_{\phi}(z_i).
\end{equation}
We store the fitted flow parameters, the log priors $\log \hat{\omega}_c$, and the empirical $Q_{0.001}$ and $Q_{0.999}$ quantiles of the resulting log-densities to normalize densities at test time.
The trained flows output class-conditional log-densities that, together with the empirical priors, yield density estimates for both seen and shifted queries.

\PAR{Query-Density Estimator Prediction.}
At inference time we evaluate the latent density of candidate queries against the stored flow ensemble.
Given a query vector $z$ from the final decoder layer, we compute the class-conditional log-densities $\log q_{\phi_c}(z)$ with the trained flows.
We then compute the marginal log-density by aggregating over classes, weighted by the empirical priors $\hat{\omega}_c$, which account for class imbalance and ensure that density estimates for rare classes are not penalized:
\begin{equation}\label{eq:11}
    \log q(z) = \log \sum_{c=1}^{C} \exp\!\left(\log q_{\phi_c}(z) + \log \hat{\omega}_c\right).
\end{equation}
The scalar $q(z)$ captures how closely the current query resembles the feature distribution of TP training objects.
Following \cite{yoon2024uncertainty}, we normalize the resulting log-density class-wise to $[0,1]$ using the cached per-class percentile statistics from the training log-densities,
\begin{equation}
    \log q(z)' = \operatorname{clip}\left(\frac{\log q(z) - Q_{0.001}}{Q_{0.999} - Q_{0.001}}\right),
\end{equation}
where $\operatorname{clip}(x) = \max(0,\min(1,x))$.
The normalized density $\log q(z)'$ is high for in-distribution queries and decreases to zero as the query drifts away from the feature manifold, as illustrated in Figure~\ref{fig:density_comparison}.

\input{figs/density_comparison_compiled}

\subsection{Density-aware Post-hoc Calibration}
Normalizing-flow densities provide a compact estimate of how familiar a query is to the training set, a signal we inject into several post-hoc calibrators.
To measure the distance of the test-time query to the calibration density, we compute the standardized log-density deviation for each query's feature vector $z$:
\begin{equation}
    z_{\text{dens}}(z) = \frac{\log q(z)' - \hat{\mu}_{\log q(z)'}}{\hat{\sigma}_{\log q(z)'}},
\end{equation}
where $\hat{\mu}_{\log q(z)'}$ and $\hat{\sigma}_{\log q(z)'}$ are the empirical mean and standard deviation of the normalized log-densities from the calibration set.

\PAR{Classification.}
Density-aware Temperature Scaling (\mbox{\textbf{DA-TS}}) retains a single temperature $T$ but modulates it per-instance using $z_{\text{dens}}$.
The instance-specific temperature becomes
\begin{equation}
    T(z) = T \cdot \phi_{\gamma}\bigl(s_T\,z_{\text{dens}}(z)\bigr),
\end{equation}
where $s_T$ is a learned scaling parameter that controls the sensitivity of the density signal and $\phi_{\gamma}(\cdot)$ is a gain function defined as $\phi_{\gamma}(x) = 1 + \gamma \tanh(x)$ which maps argument $x$ to the interval $[1-\gamma,\,1+\gamma]$.
We use $\phi_{\gamma}(\cdot)$ here for bounding the density signal to a certain range to ensure that outliers or strong shifts in the densities do not lead to overly large corrections in the temperature.
Applying the new temperature $T(z)$ to the logits $\ell$ yields the correction $\ell'(z) = \ell/T(z)$ before applying the sigmoid function.

Density-aware Platt Scaling (\textbf{DA-PS}) applies the same principle to both the scale and shift parameter
\begin{equation}
    \ell'(z) = \ell \cdot \phi_{\gamma}\bigl(s_{\text{scale}}\,z_{\text{dens}}(z)\bigr) + b \cdot \phi_{\gamma}\bigl(s_{\text{shift}}\,z_{\text{dens}}(z)\bigr).
\end{equation}

For Density-aware Isotonic Regression (\textbf{DA-IR}), we create a new feature $u(z)$ by computing a weighted combination of confidence $p(z) = \text{sigmoid}(\ell(z)) $ and $z_{\text{dens}}$, plus a bias term
\begin{equation}
    u(z) = w_s\, p(z) + w_d\,z_{\text{dens}}(z) + b.
\end{equation}
An isotonic regressor is then fitted to this new feature $u(z)$.
This is necessary since IR is nonparametric and learns a piecewise-constant mapping, and applying $z_{\text{dens}}$ directly to the mapping parameters produces unstable fits.

All calibration parameters, including the original parameters ($T$, $b$), the density-specific scaling factors (e.g., $s_T, s_{\text{scale}}$), and the feature weights ($w_s, w_d$), are jointly optimized by minimizing the NLL on the calibration set.
This ensures that the parameters are tuned to produce the most likely probability estimates, enabling the model to remain confident on high-density regions while becoming more conservative as it moves into sparser parts of the feature space.

\PAR{Regression.}
We calibrate the regression uncertainties produced by the KL-divergence uncertainty head using Density-aware Temperature Scaling (\bf{DA-TS}) applied to each predicted variance and the yaw concentration.
Let $\mathcal{S} = \left\{ \sigma_x^2,\sigma_y^2,\sigma_z^2,\sigma_{\ell}^2,\sigma_w^2,\sigma_h^2,\kappa_\theta \right\}$ denote the regression uncertainties associated with $(\hat{x},\hat{y},\hat{z},\hat{\ell},\hat{w},\hat{h},\hat{\theta})$.
For every $\sigma^2 \in \mathcal{S}$ we first apply an affine correction to the variance
\begin{equation}
    \hat{\sigma}^2 = s\,\sigma^2 + b,
\end{equation}
and then use $z_{\text{dens}}(z)$ to apply a bounded correction
\begin{equation}
    {\sigma'}^2(z) = \hat{\sigma}^2\,\,\phi_{\gamma}\!\bigl(s_{\sigma}\,z_{\text{dens}}(z)\bigr), \;\; \forall \sigma^2 \in \mathcal{S}.
\end{equation}
For each regressor, the set of coefficients $\{s,b,s_{\sigma}\}$ is learned by minimizing the MCA so that low-density queries are assigned larger uncertainties while familiar ones remain sharp.

\subsection{Baseline Methods}
\PARbegin{Sample-based.}
We use Monte-Carlo Dropout (MCD)~\cite{mcd} and Deep Ensembles (DE)~\cite{de} as baselines for both regression and classification.
Similar to \cite{pitropov2022}, we conduct repeated forward passes to produce multiple sets of detections that are clustered with DBSCAN.
We group boxes of the same class, remove singleton clusters and for each cluster we average class scores and box parameters, compute the centroid and size covariances, as well as the angular concentration.

\PAR{Classification.}
We use CalDETR~\cite{caldetr} and TCD~\cite{tcd} as \emph{train-time} calibration baselines and report implementation details and parameter choices in the \supp.
Following \cite{kuzucu2024calibration}, we implement Platt Scaling (PS)~\cite{ps}, Isotonic Regression (IR)~\cite{ir} and additionally Temperature Scaling (TS)~\cite{ts} as \emph{post-hoc} calibration baselines.

\PAR{Regression.}
To introduce a simple regression baseline, we use an affine depth-based calibrator (Depth) that models the uncertainty as a linear function of distance to the object. 
Given the object's distance $d$, the variance for each regression parameter is modeled as $\sigma^2(d) = a_{\sigma^2} d + b_{\sigma^2}$ (angular concentration: $\kappa(d) = a_{\kappa} d + b_{\kappa}$), where $a_{\kappa}$ and $b_{\kappa}$ are learned parameters for each regression output. 
Additionally, we follow the CalDETR~\cite{caldetr} pipeline, and interpret the variance along the transformer layers as a measure of uncertainty.
For each query, we collect the bounding box predictions from all six layers and compute their empirical variance.
Following~\cite{trust}, we implement Temperature Scaling for each regression parameter ${\sigma'}^2 \leftarrow \sigma^2 / T_{\sigma^2}, \; \forall \; \sigma^2 \in \left\{ \sigma_x^2,\sigma_y^2,\sigma_z^2,\sigma_{\ell}^2,\sigma_w^2,\sigma_h^2,\kappa_\theta \right\}$ and minimize the respective MCA on the calibration set.
We also report results obtained with an energy-based loss~\cite{es}, as an alternative to KL~\cite{kld}.

\PAR{Accounting for Class Imbalance.}
Each density-aware calibrator and its naïve counterpart is trained in a \bf{global} fashion (\gbox), treating all classes as one class, and a \bf{class-wise} variant (\cbox), training a separate calibrator for each class.

%% file: figs/method_overview.tex
\begin{figure*}[tp]
    \centering
    \includegraphics[width=\linewidth]{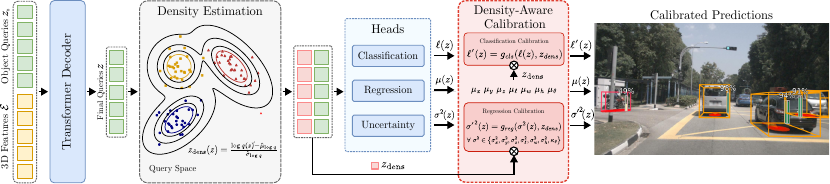}
    \caption{
        \textbf{Overview of Query2Uncertainty}
        We feed 3D features (either LiDAR or multi-view camera) as tokens into a standard DETR-style 3D object detector where they interact with object queries $z$.
        The refined queries at the last decoder layer are passed through a feature density estimator, that returns $z_{dense}$ a measure how well each individual query aligns with True Positives queries from the train set.
        This module is followed by the detection heads that return probabilistic 3D box parameters and class scores.
        To obtain well calibrated uncertainties for ID and distribution shift data, we propose our density-aware post-hoc calibration module that calibrates box variances and scores.
    }
    \label{fig:method_overview}
\end{figure*}

%% file: figs/density_comparison_compiled.tex
\begin{figure}[t]
    \centering
    \includegraphics[width=0.47\textwidth, trim={0.325cm 0.3cm 0 0}]{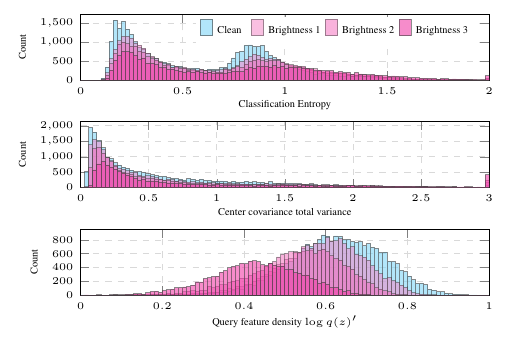}
    \caption{
        \textbf{Uncertainties under distributional shift.}
        We visualize classification entropy, centroid total variance, and query density for increasing brightness severity levels. The progressive drift of the query density histogram indicates that the density estimator effectively captures distributional shifts in the input data.
    }
    \label{fig:density_comparison}
\end{figure}

%% file: 04_experiments.tex
\section{Experiments}
We evaluate the calibration performance of our density-based methods using the nuScenes dataset~\cite{caesar2020nuscenes} for In-Distribution (ID) data and MultiCorrupt~\cite{beemelmanns2024multicorrupt} for distribution-shift scenarios.
Our approach outperforms sample-based methods, classic post-hoc methods, and other baselines in both ID and distribution-shift scenarios.

\subsection{Experimental Setup}
The nuScenes dataset~\cite{caesar2020nuscenes} consists of a diverse set of urban scenarios across multiple cities, where each frame includes six multi-view camera images, LiDAR and radar data.
We train our camera and LiDAR-based detectors on the train split and evaluate on the validation split for ID accuracy performance.
Both models are trained with 900 queries, for 32 epochs, following the training protocol of PETR~\cite{liu2022petr}.
To calibrate our probabilistic detectors, we follow~\cite{kuzucu2024calibration} and split the validation set sequentially into a calibration set (40\%) and a test set (60\%).
For distribution-shifted tests, we use MultiCorrupt~\cite{beemelmanns2024multicorrupt}, which contains ten different LiDAR and camera corruptions, at three severity levels, applied to the nuScenes validation set.
We tune all post-hoc calibrators on the calibration set using Differential Evolution~\cite{storn1997differential} with 20,000 iterations, which yields more deterministic and accurate results compared to gradient-based optimizers.
During calibration we also store the mean and variance of the normalized query densities for each class to measure the deviation from the ID query densities at test time.
For DE we use six ensemble members, each trained with different a random seed, and for MCD we use six forward passes with a dropout rate of $0.3$ during inference.
In contrast to~\cite{kuzucu2024calibration}, where only a single threshold is used for evaluation, we evaluate the quality of the uncertainty estimates at different confidence thresholds from $0.05$ to $0.60$ with increments of $0.05$, and report the mean over all thresholds.
We limit the threshold at $0.60$ to avoid missing predictions for rare classes%

\subsection{In-Distribution Evaluation}

\input{tables/0_main_cls_mean.tex}
\input{tables/0_main_area_reg_mean.tex}

We report the threshold-averaged classification and regression calibration errors in Table~\ref{tab:0_main_cls_mean} and Table~\ref{tab:0_main_area_reg_mean}, respectively.
DE and MCD perform poorly on both tasks, and also suffer a drop in accuracy.
We ablated different clustering and merging strategies, size of the ensemble and MCD samples (see \supp), but the performance remained at the level of an uncalibrated baseline.
Previous work~\cite{pitropov2022} focuses on the KITTI dataset, which contains only three classes, whereas nuScenes comprises ten categories, with object-dense scenes, including many small objects such as traffic cones and barriers whose boxes are often grouped closely together.
This causes the clustering and merging to over-group adjacent boxes, making it difficult to obtain reliable consensus detections.
Both training-time methods, CalDETR~\cite{caldetr} and TCD~\cite{tcd}, also struggle to improve calibration performance.
Since both approaches are designed for 2D object detection, they do not translate well to the 3D detection task.
TCD is IoU-based, and the intersection in 3D decreases sharply compared to the 2D case.
This leads small objects to have little or zero IoU with the ground truth boxes, make the loss signal noisy and unreliable.
CalDETR, originally designed for 2D detection using COCO, is not designed to consider large class imbalance present in nuScenes.
For \emph{classification}, our proposed density-aware calibrators outperform their non-density-aware counterparts in almost all cases, indicating that density deviations provide useful information for refining ID-trained post-hoc calibrators.
For \emph{regression}, the uncertainty head trained with KL loss~\cite{kld} provides a strong baseline, but requires calibration to minimize Miscalibration Area (MCA).
Our DA-TS method improves over the TS baseline in almost all cases, achieving the best regression calibration performance across both detectors.
Figures~\ref{fig:petr_reliability_diagrams} and \ref{fig:petr_mca_diagrams} illustrate example reliability diagrams for classification and regression, respectively. 
For classification, both the uncalibrated model and DE are overconfident in their predictions, which is reduced by the post-hoc calibrators, bringing the empirical confidences close to the ideal calibration line.
For regression, the uncalibrated variances of the KL uncertainty head exhibit miscalibration in both directions, underconfidence and overconfidence.
In both cases, density-aware calibration aligns predicted with empirical coverage.
These results further indicate that, on imbalanced datasets, post-hoc methods benefit from class-wise calibration (\cbox).

\subsection{Distribution-Shift Evaluation}
We use MultiCorrupt~\cite{beemelmanns2024multicorrupt} to evaluate the robustness of uncertainty estimates under distribution-shift and report the mean calibration performance across all corruptions and their three severity levels in Table~\ref{tab:2_multicorrupt_cls_mean} and Table~\ref{tab:2_multicorrupt_reg_mean}.
The calibration quality for the post-hoc methods drops under corruptions and is close to that of the uncalibrated model.
However, in both classification and regression, our density-aware calibrators consistently outperform their non-density-aware counterparts, demonstrating the effectiveness of density-based adjustments for handling distribution shift.
To evaluate robustness against semantic distribution shifts, we conduct a transfer experiment using the \texttt{Boston} (train \& calib.) and \texttt{Singapore} (test) scenes of nuScenes.
As shown in Table~\ref{tab:b2s_cls}, our density-aware methods consistently improves classification calibration.
For regression, where geometric features remain relatively invariant across locations, our method maintains robust performance (Table~\ref{tab:b2s_reg}).
\input{tables/2_multicorrupt_cls_mean.tex}
\input{tables/2_multicorrupt_reg_mean.tex}
\input{tables/2_boston_singapore.tex}

\subsection{Density Estimator Comparison}
The choice of density estimator can impact the quality of the density estimates and thus the performance of our density-aware calibrators.
We compare a Gaussian Mixture Model (GMM)~\cite{ddu} and Normalizing Flows (NF)~\cite{dinh2017density} as density estimators and report the results for the distribution-shifted setting in Table~\ref{tab:ablation_gda_domain}.
\input{tables/1_ablation_gda.tex}
NF slightly outperforms GMM, indicating that NF is better suited for capturing complex latent feature distributions, albeit at the cost of an increased number of parameters and higher latency.

\vspace{-4pt}

%% file: tables/0_main_cls_mean.tex
\begin{table*}[h]
\begin{minipage}[t]{12.50cm}
\centering
\caption{
\textbf{In-Distribution - Classification Calibration}
D-ECE and LaECE (in \%) are reported as mean over detections thresholds from $[0.05, 0.60]$ with 0.05 steps.
Class-wise calibration: \cbox; Global calibration: \gbox.
\textcolor{green!60!black}{+}/\textcolor{red!65!black}{-}: better/worse than the naïve post-hoc method.
}
\adjustbox{max width=12.30cm}{%
\begin{tabular}{w{c}{1.0cm}||c|w{c}{1.1cm}w{c}{1.1cm}|cc||w{c}{1.1cm}w{c}{1.1cm}|cc}
\toprule
\rowcolor{black!10!white} & & \multicolumn{4}{c||}{PETR} & \multicolumn{4}{c}{SECOND} \\
\midrule
Calib. & Calib. &  \multicolumn{2}{c|}{Calibration Error}     & \multicolumn{2}{c||}{Accuracy} & \multicolumn{2}{c|}{Calibration Error}  & \multicolumn{2}{c}{Accuracy} \\
Type   & Method & D-ECE$\downarrow$ & LaECE$\downarrow$ &  mAP$\uparrow$ & NDS$\uparrow$ & D-ECE$\downarrow$  &  LaECE$\downarrow$  &  mAP$\uparrow$  & NDS$\uparrow$ \\
\midrule
None & Uncal.   &             8.556 &            27.211 &          38.25 &        45.05  &           15.509  &            17.378 & 54.53 & 63.57 \\
\midrule
\multirow{2}{*}{\shortstack{Sample\\-based}} 
 & MCD~\cite{mcd}         &         9.487 &      30.110 &          36.90 &        44.23  &            16.517 &            19.523 & 51.80 & 62.30 \\
 & DE~\cite{de}           &         9.091 &      32.754 &          33.20 &        43.39  &            15.997 &            19.635 & 50.83 & 62.36 \\
\midrule
\multirow{2}{*}{\shortstack{Train\\-time}}
 & CalDETR~\cite{caldetr} &         9.264 &      30.089 &          37.12 &         44.28  &           15.450 &            19.906 & 53.57 & 63.17 \\
 & TCD~\cite{tcd}         &         9.567 &      24.533 &          38.27 &         45.05  &           18.182 &            18.869 & 54.37 & 63.81 \\
\midrule
\multirow{6}{*}{\shortstack{Post\\-hoc}}
 & TS~\cite{ts} \gbox     &         7.859 &      26.115 &          38.26 &         45.04  &           10.050 &           16.194  & 54.53 & 63.58 \\
 & TS~\cite{ts} \cbox     &         5.312 &      24.660 &          38.26 &         45.04  &            7.818 &           13.968  & 54.53 & 63.58 \\
 & PS~\cite{ps} \gbox     &         7.752 &      25.686 &          38.26 &         45.04  &           10.092 &           15.810  & 54.53 & 63.58 \\
 & PS~\cite{ps} \cbox     &         3.996 &      22.782 &          38.26 &         45.04  &            2.311 &            9.076  & 54.53 & 63.58 \\
 & IR~\cite{ir} \gbox &            7.451 &       25.753 &          38.26 &         45.04  &            9.900 &           15.642  & 54.53 & 63.58 \\
 & IR~\cite{ir} \cbox &       \un{2.999} &       22.869 &          38.26 &         45.04  &       \un{1.867} &            9.078  & 54.53 & 63.58 \\
\midrule
 \multirow{6}{*}{\shortstack{Density\\-aware\\(Ours)}}
 & DA-TS \gbox    & 7.688\da{2.18}      & 24.831\da{4.92}      & 38.36 & 45.03  & 9.676\da{3.72}      & 15.185\da{6.23}      & 54.36 & 63.65 \\
 & DA-TS \cbox    & 5.772\da{-8.66}     & 24.185\da{1.93}      & 38.36 & 45.03  & 7.351\da{5.97}      & 13.237\da{5.23}      & 54.36 & 63.65 \\
 & DA-PS \gbox    & 7.598\da{1.99}      & 24.960\da{2.83}      & 38.36 & 45.03  & 9.628\da{4.60}      & 15.043\da{4.85}      & 54.36 & 63.65 \\
 & \cb DA-PS \cbox    & \cb 3.899\da{2.43}      & \cb \bf{22.393}\da{1.71} & \cb 38.36 & \cb 45.03  & \cb 2.224\da{3.76}      & \cb \bf{8.812}\da{2.91}  & \cb 54.36 & \cb 63.65 \\
 & DA-IR \gbox    & 7.209\da{3.25}      & 24.015\da{6.75}      & 38.36 & 45.03   &  5.217\da{47.30}   &  10.982\da{29.79}    & 54.36 & 63.65 \\
 & \cb DA-IR \cbox    & \cb \bf{2.955}\da{1.47} & \cb \un{22.550}\da{1.39} & \cb 38.36 & \cb 45.03  & \cb \bf{1.839}\da{1.50} & \cb \un{8.877}\da{2.21}  & \cb 54.36 & \cb 63.65 \\
\bottomrule
\end{tabular}%
}%
\label{tab:0_main_cls_mean}
\end{minipage}\hfill
\begin{minipage}[t]{4.50cm}
\centering
\captionof{figure}{
\textbf{Example Reliability Diagram}
(\mbox{D-ECE}) for SECOND.
}
\label{fig:petr_reliability_diagrams}
\vspace{8pt}
\begin{subfigure}[t]{4.40cm}
\centering
\begin{tikzpicture}
\node[inner sep=0pt] (img) {\includegraphics[width=3.25cm,trim=0 1cm 0 0,clip]{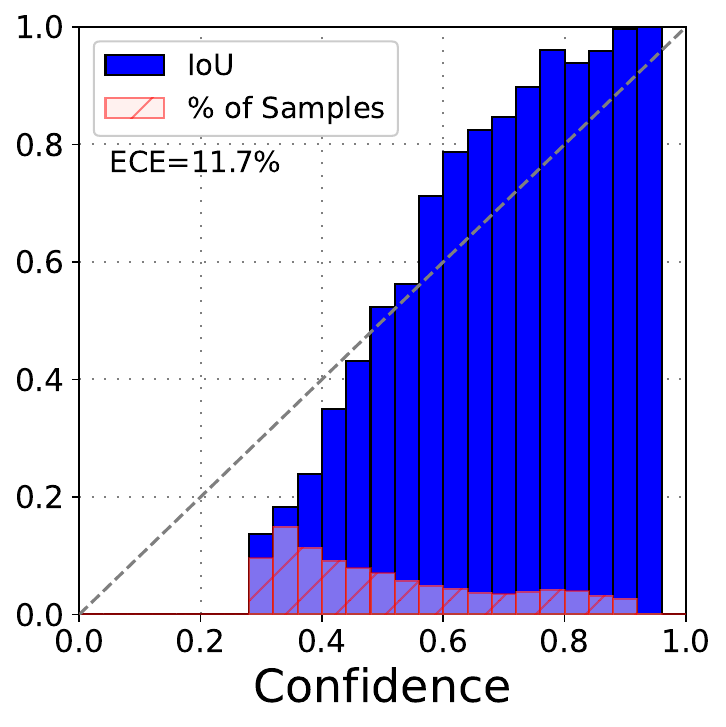}};
\node[rotate=90, anchor=south, font=\scriptsize, xshift=0.1cm, yshift=-0.1cm] at (img.west) {Accuracy};
\node[font=\scriptsize, xshift=0.0cm, yshift=-0.1cm] at (img.south) {Confidence};
\end{tikzpicture}
\vspace{-3pt}
\caption{Uncalibrated}
\label{fig:rd_uncal}
\end{subfigure}\\[0pt]
\begin{subfigure}[t]{4.40cm}
\centering
\begin{tikzpicture}
\node[inner sep=0pt] (img) {\includegraphics[width=3.25cm,trim=0 1cm 0 0,clip]{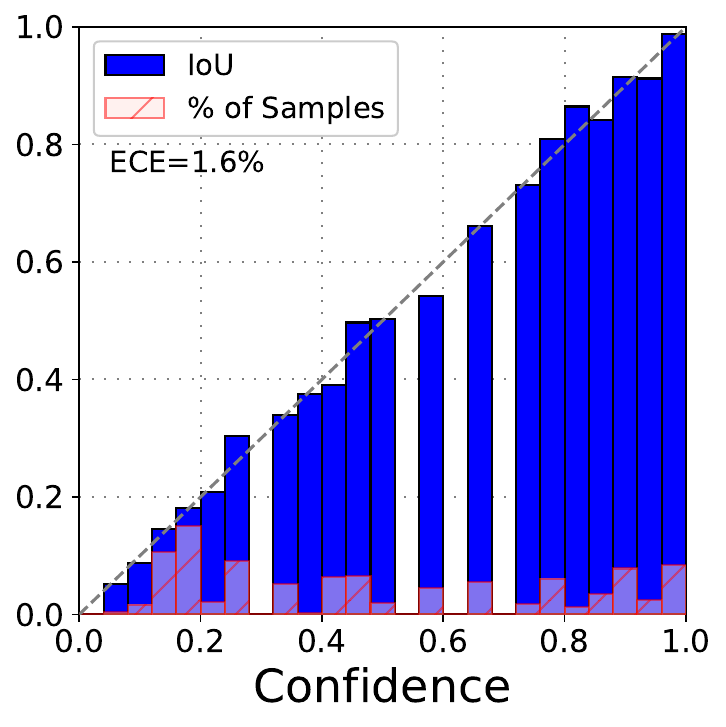}};
\node[rotate=90, anchor=south, font=\scriptsize, xshift=0.1cm, yshift=-0.1cm] at (img.west) {Accuracy};
\node[font=\scriptsize, xshift=0.0cm, yshift=-0.1cm] at (img.south) {Confidence};
\end{tikzpicture}
\vspace{-3pt}
\caption{Calibrated by DA-IR}
\label{fig:rd_daps}
\end{subfigure}
\end{minipage}
\end{table*}

%% file: tables/0_main_area_reg_mean.tex
\begin{table*}[h]
\begin{minipage}[t]{13.2cm}
\centering
\caption{
\textbf{In-Distribution - Regression Calibration}
Metrics (MCA for centroid, size and orientation in \%) are reported as mean over detections thresholds from $[0.05, 0.60]$ with 0.05 steps. 
Class-wise calibration:~\cbox; Global calibration:~\gbox.
\textcolor{green!60!black}{+}/\textcolor{red!65!black}{-}: better/worse than the naïve post-hoc method.
}
\adjustbox{max width=13cm}{%
\begin{tabular}{w{c}{0.75cm}||w{c}{1.75cm}|w{c}{1.1cm}w{c}{1.1cm}w{c}{1.1cm}|w{c}{1.00cm}w{c}{1.00cm}||w{c}{1.1cm}w{c}{1.1cm}w{c}{1.1cm}|w{c}{1.00cm}w{c}{1.00cm}}
\toprule
\rowcolor{black!10!white} & & \multicolumn{5}{c||}{PETR} & \multicolumn{5}{c}{SECOND} \\
\midrule
Calib. &  Calib. &  \multicolumn{3}{c|}{Calibration Error} & \multicolumn{2}{c||}{Accuracy} & \multicolumn{3}{c|}{Calibration Error} & \multicolumn{2}{c}{Accuracy} \\
Type & Method & $\text{MCA}_{xyz}\!\!\downarrow$& $\text{MCA}_{lwh}\!\!\downarrow$ & $\text{MCA}_{\phi}\!\!\downarrow$ & mAP$\uparrow$ & NDS$\uparrow$ & $\text{MCA}_{xyz}\!\!\downarrow$ & $\text{MCA}_{lwh}\!\!\downarrow$ & $\text{MCA}_{\phi}\!\!\downarrow$ & mAP$\uparrow$ & NDS$\uparrow$ \\
\midrule
\multirow{3}{*}{\shortstack{Sample\\-based}}
 & MCD~\cite{mcd}         &  35.878  & 42.135 & 34.356           & 36.90 & 44.23      & 31.620 & 40.353 & 32.627 & 51.80 & 62.30 \\
 & DE~\cite{de}           &  28.871  & 38.092 & 27.257           & 33.20 & 43.39      & 23.089 & 39.337 & 31.997 & 50.83 & 62.36 \\
 & CalDETR~\cite{caldetr} &  31.834  & 39.713 & 27.551           & 37.12 & 44.28      & 34.234 & 38.844 & 29.500 & 53.69 & 63.17 \\
\midrule
\multirow{2}{*}{\shortstack{Train\\-time}}
 & ES~\cite{es}           &  6.930  & 7.103   & 11.49            & 38.27 & 45.03      & 3.400 & 6.235 &  8.140 & 54.53 & 63.57 \\
 & KL~\cite{kld}          &  4.384  & 4.620   & 11.39            & 38.27 & 45.05      & 4.692 & 4.627 & 14.631 & 54.53 & 63.57 \\
\midrule
\multirow{4}{*}{\shortstack{Post\\-hoc}}
 & Depth \gbox           &  8.254      & 15.381    &  18.469     & 38.26 & 45.04      &     12.494 &      13.884 &     16.741 & 54.53 & 63.57 \\
 & Depth \cbox           &  2.407      & 2.442     &  9.713      & 38.26 & 45.04      &      4.749 &      2.112  &     7.683  & 54.53 & 63.57 \\
 & TS~\cite{trust} \gbox & 4.028       & 5.106     &  9.657      & 38.26 & 45.04      &      3.443 &      5.503  &      7.521 & 54.53 & 63.57 \\
 & TS~\cite{trust} \cbox & \un{1.692}  & \bf{2.139}&  \un{8.102} & 38.26 & 45.04      & \un{1.533} & \un{1.919}  & \un{6.040} & 54.53 & 63.57 \\
\midrule
\multirow{2}{*}{\shortstack{Density\\-aware}}
 & DA-TS \gbox           & 4.059\da{-0.77}     & 5.236\da{-2.55} &      9.710\da{-0.55} & 38.36 & 45.03 &      2.875\da{16.50} &      4.693\da{14.72} &       7.309\da{2.82} & 54.36 & 63.65 \\
 & \cb DA-TS \cbox        & \cb \bf{1.538}\da{9.10} & \cb \un{2.141}\da{-0.09} & \cb \bf{8.099}\da{0.04}  & \cb 38.36 & \cb 45.03 & \cb \bf{1.518}\da{0.98} & \cb \bf{1.758}\da{8.39} & \cb \bf{6.037}\da{0.05}  & \cb 54.36 & \cb 63.65 \\
\bottomrule
\end{tabular}%
}%
\label{tab:0_main_area_reg_mean}
\end{minipage}\hfill
\begin{minipage}[t]{3.8cm}
\centering
\captionof{figure}{
\textbf{Example Miscal. Area}
($\text{MCA}_{xyz}$) for PETR.
}
\label{fig:petr_mca_diagrams}
\vspace{-2pt}
\begin{subfigure}[t]{2.75cm}
\centering
\includegraphics[width=2.40cm]{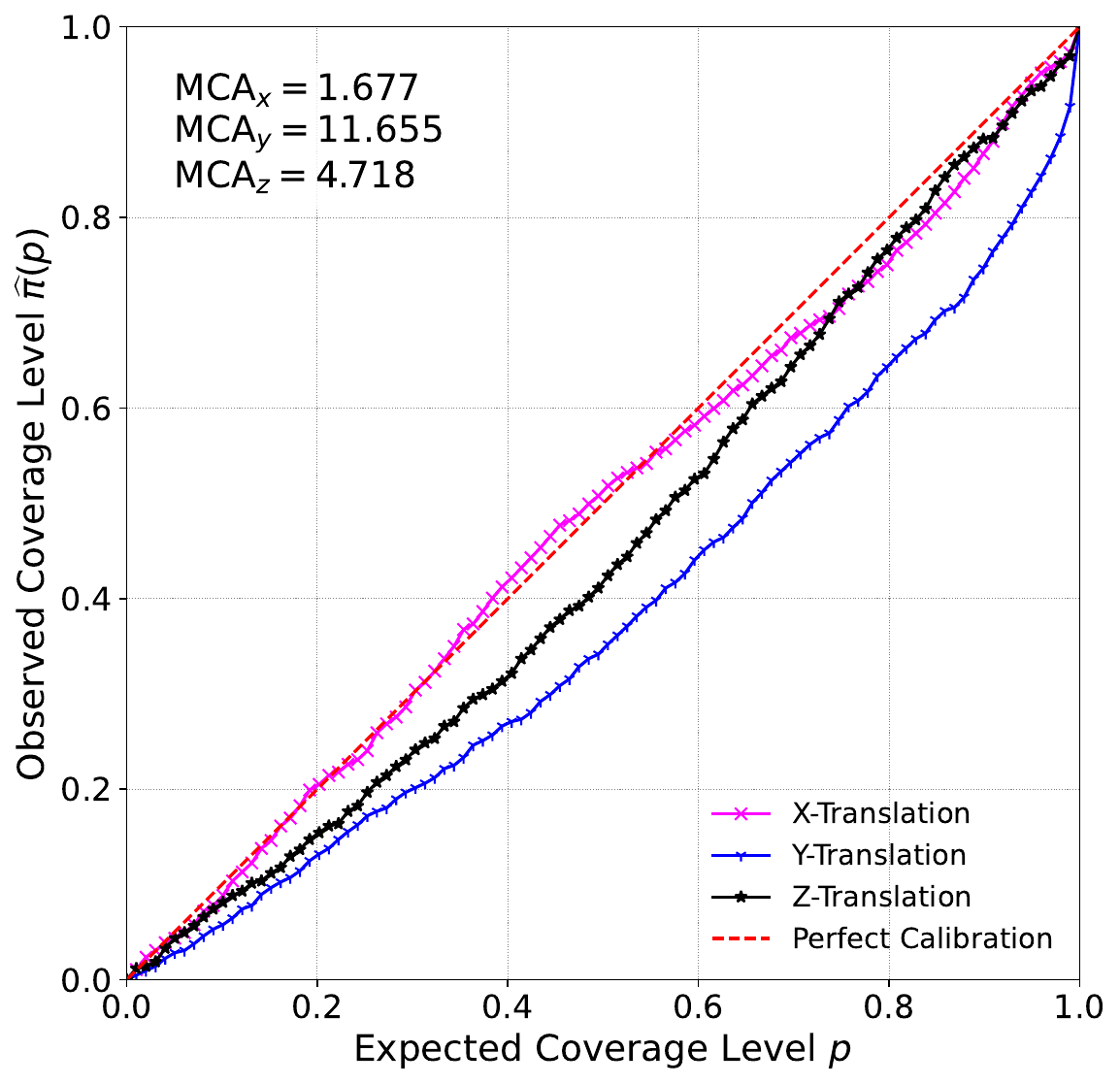}
\vspace{-5pt}
\caption{Uncalibrated KL~\cite{kld}}
\label{fig:mca_uncal}
\end{subfigure}\\[-2pt]
\begin{subfigure}[t]{2.75cm}
\centering
\includegraphics[width=2.40cm]{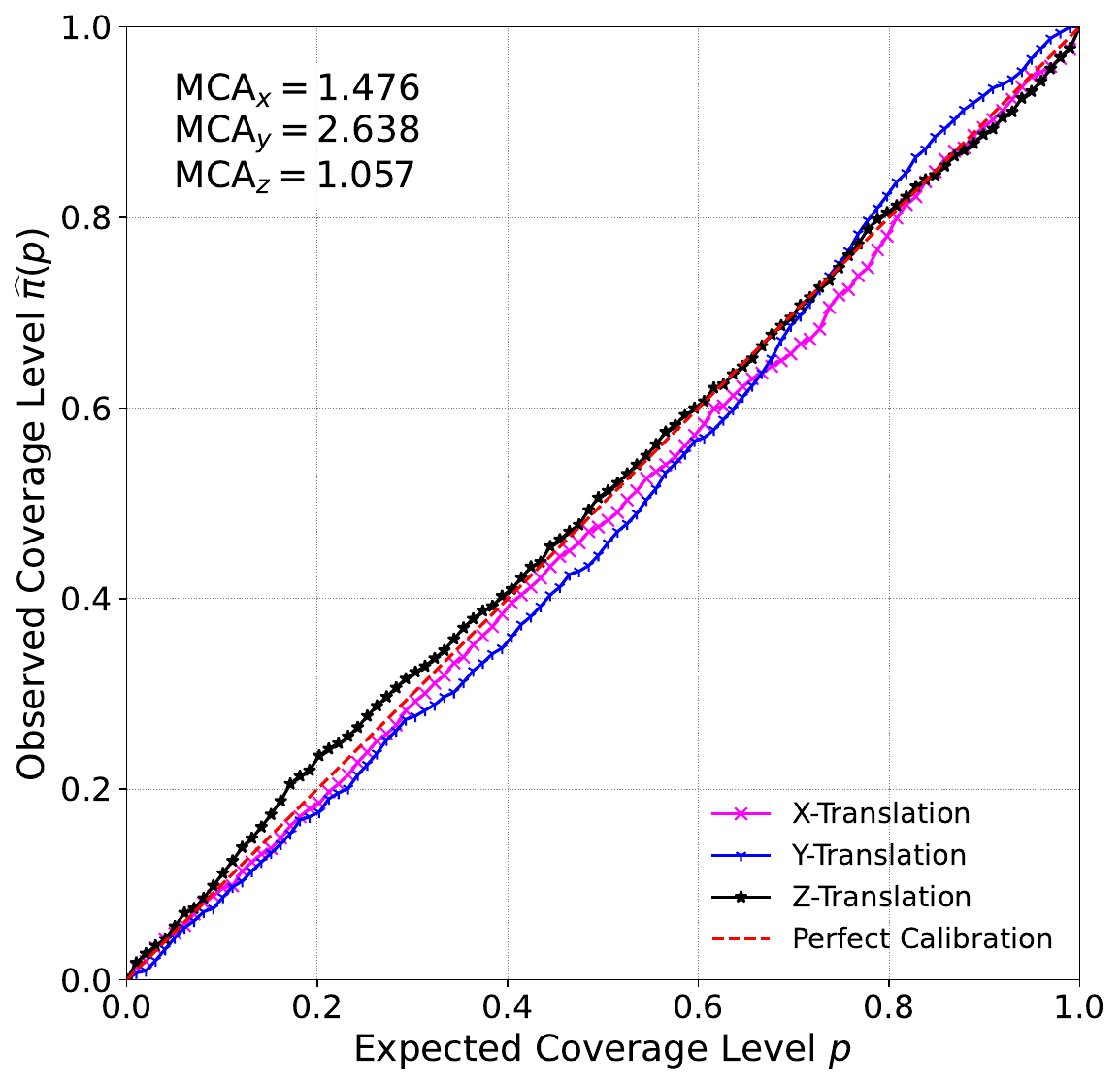}
\vspace{-5pt}
\caption{Calibrated by DA-TS}
\label{fig:mca_dats}
\end{subfigure}
\end{minipage}
\end{table*}

%% file: tables/2_multicorrupt_cls_mean.tex
\begin{table}[t]
\centering
\caption{
\textbf{Distribution Shift - Classification Calibration}
Class-wise calibration: \cbox; Global calibration: \gbox. \textcolor{green!60!black}{+}/\textcolor{red!65!black}{-}: better/worse than the naïve post-hoc method.
}
\adjustbox{max width=\columnwidth}{%
\begin{tabular}{c||c|c|c||c|c}
\toprule
\rowcolor{black!10!white} & & \multicolumn{2}{c||}{PETR} & \multicolumn{2}{c}{SECOND} \\
\midrule
Calib. & Calib. & \multirow{2}{*}{D-ECE$\downarrow$} & \multirow{2}{*}{LaECE$\downarrow$} & \multirow{2}{*}{D-ECE$\downarrow$} & \multirow{2}{*}{LaECE$\downarrow$} \\
Type & Method   &                                    &                                    &                                    &                                    \\
\midrule
None & Uncal.   &                             13.040 &                            29.059  &                              22.490 &                            21.558  \\
\midrule
\multirow{6}{*}{\shortstack{Post\\-hoc}} 
 & TS~\cite{ts} \gbox &          12.271 &         27.671  &          15.758 &      19.638   \\
 & TS~\cite{ts} \cbox &          10.271 &         26.881  &          13.363 &      17.319   \\
 & PS~\cite{ps} \gbox &          12.147 &         27.510  &          15.812 &      19.523   \\
 & PS~\cite{ps} \cbox &           7.990 &         24.602  &           7.037 &      11.052   \\
 & IR~\cite{ir} \gbox &          11.969 &         27.429  &          15.689 &      19.441   \\
 & IR~\cite{ir} \cbox &        \un{7.449} &         24.333  &           7.076 &      11.093   \\
\midrule
\multirow{6}{*}{\shortstack{Density\\-aware\\(Ours)}}
 & DA-TS \gbox        &     11.382\da{7.24} &      25.825\da{6.67} &     15.257\da{3.18} &     18.657\da{5.00} \\
 & DA-TS \cbox        &      9.655\da{6.00} &      25.649\da{4.58} &     13.018\da{2.58} &     16.364\da{5.51} \\
 & DA-PS \gbox        &     11.396\da{6.18} &      26.308\da{4.37} &     15.368\da{2.81} &     18.556\da{4.95} \\
 & \cb DA-PS \cbox        &    \cb  7.770\da{2.75} & \cb \un{23.131}\da{5.98} &  \cb \bf{6.777}\da{3.69} & \cb 10.051\da{9.06} \\
 & DA-IR \gbox        &     10.802\da{9.75} &      24.315\da{11.35} &      9.966\da{36.48} &     11.879\da{38.90} \\
 & \cb DA-IR \cbox     & \cb \bf{7.211}\da{3.20} & \cb \bf{22.212}\da{8.72} & \cb \un{6.895}\da{2.56} & \cb \bf{9.273}\da{16.41} \\
\bottomrule
\end{tabular}%
}%
\label{tab:2_multicorrupt_cls_mean}
\end{table}

%% file: tables/2_multicorrupt_reg_mean.tex
\begin{table}[t]
\centering
\caption{
\textbf{Distribution Shift - Regression Calibration}
Class-wise calibration: \cbox; Global calibration: \gbox. \textcolor{green!60!black}{+}/\textcolor{red!65!black}{-}: better/worse than the naïve post-hoc method.
}
\adjustbox{max width=\linewidth}{%
\begin{tabular}{w{c}{0.80cm}||w{c}{1.5cm}|c|c|c||c|c|c}
\toprule
\rowcolor{black!10!white} & & \multicolumn{3}{c||}{PETR} & \multicolumn{3}{c}{SECOND} \\
\midrule
Calib. &  Calib. & \multirow{2}{*}{$\text{MCA}_{xyz}\!$} & \multirow{2}{*}{$\text{MCA}_{lwh}\!$} & \multirow{2}{*}{$\text{MCA}_{\phi}\!$} & \multirow{2}{*}{$\text{MCA}_{xyz}\!$} & \multirow{2}{*}{$\text{MCA}_{lwh}\!$} & \multirow{2}{*}{$\text{MCA}_{\phi}\!$} \\
Type & Method &  &  &  &  &  \\
\midrule
\multirow{2}{*}{\shortstack{Train\\-time}}
 & ES~\cite{es}            & 8.187            &             6.730  &          12.713 &             6.085 &      6.152  & 15.103 \\
 & KL~\cite{kld}           & 5.101            &             6.233  &          12.500 &             6.363 &      6.332  & 16.328 \\
\midrule
\multirow{4}{*}{\shortstack{Post\\-hoc}}
 & Depth \gbox             & 12.742       &             16.763 &          20.830  &         16.716  &       16.198 &     16.182 \\
 & Depth \cbox             & 7.367        &             4.674  &          12.236  &         9.897   &       6.111  &      8.781  \\
 & TS~\cite{trust} \gbox   & 4.970        &             6.050  &          12.370  &         5.686   &       7.025  &      14.338 \\
 & TS~\cite{trust} \cbox   & \un{3.780}   &         \un{4.226} &      \un{10.291} &     \un{5.264} &    \un{5.096} &  \un{8.379}  \\
\midrule
\multirow{2}{*}{\shortstack{Density\\-aware}}
 & DA-TS \gbox             &      4.735\da{4.73} &       6.910\da{-14.21} &     10.486\da{15.23} &      4.605\da{19.01} &       5.863\da{16.54} &      9.207\da{35.79} \\
 & \cb DA-TS \cbox             & \cb \bf{3.546}\da{6.19} &  \cb \bf{4.217}\da{0.21} & \cb \bf{9.533}\da{7.37} & \cb \bf{4.037}\da{23.31} & \cb \bf{3.941}\da{22.66} & \cb \bf{8.105}\da{3.27} \\
\bottomrule
\end{tabular}%
}%
\label{tab:2_multicorrupt_reg_mean}
\end{table}

%% file: tables/2_boston_singapore.tex
\begin{table}[t]
\caption{\bf{Semantic Shift - Classification - Boston $\to$ Singapore}
Using PETR  and class-wise calibration (\cbox) for all results.
}
\label{tab:b2s_cls}
\centering
\resizebox{\columnwidth}{!}{%
\begin{tabular}{l|c|ccc|ccc}
\toprule
\rowcolor{black!10!white} Metric & Uncal. & TS & PS & IR & DA-TS & DA-PS & DA-IR \\
\midrule
D-ECE$\downarrow$  & 10.109 & 8.510  & 6.340  & 5.137  & 7.987\da{1}  & 5.789\da{1}  & \cb\bf{3.645}\da{1} \\
LaECE$\downarrow$  & 26.682 & 25.747 & 24.880 & 24.731 & 25.055\da{1} & 23.354\da{1} & \cb\bf{21.810}\da{1} \\
\bottomrule
\end{tabular}
}
\vspace{-0.05in}
\end{table}
\begin{table}[t]
\scriptsize
\centering
\caption{
\bf{Semantic Shift - Regression - Boston $\to$ Singapore}
Using PETR and class-wise calibration (\cbox) for all results.
}
\label{tab:b2s_reg}
\begin{tabular}{l|c|cc}
\toprule
\rowcolor{black!10!white} Metric & Uncal. & TS & DA-TS \\
\midrule
$\text{MCA}_{xyz}\!\!\downarrow$ & 10.105 & 8.584 & \cb\bf{8.442}\da{1} \\
$\text{MCA}_{lwh}\!\!\downarrow$ & 11.793 & 7.468 & \cb\bf{7.360}\da{1} \\
$\text{MCA}_{\phi}\!\!\downarrow$  & 10.778 & 9.771 & \cb\bf{9.160}\da{1} \\
\bottomrule
\end{tabular}
\vspace{-0.05in}
\end{table}

%% file: tables/1_ablation_gda.tex
\begin{table}[t]
\caption{
\bf{Distribution Shift - Density Estimator Comparison}
Using PETR and class-wise calibration (\cbox).
Density estimator latency measured on an NVIDIA A100 with a batch size of 300.
}
\centering
\resizebox{\columnwidth}{!}{%
\begin{tabular}{cccccccc}
\toprule
\rowcolor{black!10!white} Model &  Estimator     & D-ECE$\downarrow$ & $\text{MCA}_{xyz}\!\!\downarrow$ & $\text{MCA}_{lwh}\!\!\downarrow$ & $\text{MCA}_{\phi}\!\!\downarrow$ & Params$\downarrow$ & Latency$\downarrow$ \\
\midrule
\multirow{2}{*}{\shortstack{PETR}}
      & GMM~\cite{ddu}            &        7.430 &           3.612   &  \bf{4.206}  &        9.542  & \bf{0.7M} & \bf{0.58}~ms \\
      & NF~\cite{dinh2017density} &   \bf{7.211} &       \bf{3.546}  &      4.217   &    \bf{9.533} &     2.1M  &     8.99~ms  \\
 \cmidrule{2-8}
\multirow{2}{*}{\shortstack{SEC-\\OND}}
      & GMM~\cite{ddu}            &        6.997 &           4.085   &       4.053  &        8.583  & \bf{0.7M} & \bf{0.58}~ms \\
      & NF~\cite{dinh2017density} &   \bf{6.777} &       \bf{4.037}  &  \bf{3.941}  &    \bf{8.105} &    2.1M   &     8.99~ms  \\
\bottomrule
\end{tabular}
}
\vspace{-0.05in}
\label{tab:ablation_gda_domain}
\end{table}

%% file: 05_conclusion.tex
\section{Conclusion}\label{sec:conclusion}

We proposed a novel density-aware post-hoc calibration method that effectively couples feature density estimation with common calibration methods.
Our approach successfully refines both classification and regression uncertainty estimates for modern query-based 3D detectors.
Through extensive experiments, we demonstrate that our method outperforms strong baselines on both multi-view camera and LiDAR-based models.
Notably, this performance holds true not only for in-distribution data but also across challenging distribution shift.
To enable this and future research, we established a comprehensive benchmark adapting the metrics for evaluating the quality of uncertainty quantification in 3D object detection for both classification and bounding box regression.
Our findings suggest that feature density is a powerful signal for improving uncertainty calibration in autonomous driving.

\section*{Acknowledgements}
This work has received funding from the European Union’s Horizon Europe Research and Innovation Programme under Grant Agreement No. 101076754 - AIthena project.
A. Nekrasov acknowledges funding by BMBF project ``WestAI'' (grantwno. 01IS22094D).
Computations were performed with computing resources granted by RWTH Aachen University under project \texttt{rwth1901}.

%% file: 07_appendix.tex
\section{Appendix Section}\label{sec:appendix_section}

\makeatletter
\newcommand{\appendixtableofcontents}{%
  \section*{Appendix Contents}%
  \@starttoc{aptoc}}
\makeatother

\appendixtableofcontents

\subsection{Uncertainty Evaluation Framework}
\addcontentsline{aptoc}{subsection}{\protect\numberline{\thesubsection}Uncertainty Evaluation Framework}

\PAR{Overview.}
Our uncertainty evaluation framework builds upon the official \texttt{nuScenes-devkit}~\cite{caesar2020nuscenes} and extends existing data structures and evaluation routines, which makes our implementation easy to use and integrate into existing 3D object detection pipelines.
In particular, we extend the existing 3D box object class with additional regression uncertainty measures and introduce metrics that ingest classification confidences and 3D box covariance estimates while reusing the nuScenes ground-truth matching and metric reporting.

\PAR{Interface.}
As a basis for our implementation, we extend the class \texttt{NuScenesBox} with \texttt{NuScenesBoxUQ} (see Listing~\ref{lst:nuscenes_box_uq}) to include variances for centroid and size, and additionally an angular concentration for the rotation.

\begin{lstlisting}[language=Python, caption={Interface for NuScenes box structure with additional uncertainty fields for centroid, size and orientation.}, label=lst:nuscenes_box_uq]
class NuScenesBoxUQ(NuScenesBox):
  def __init__(
    self,
    center: List[float],
    center_cov: List[float],           <- Added
    bbox_list: List[List[float]],
    size: List[float],
    size_cov: List[float],             <- Added
    orientation: Quaternion,
    orientation_kappa: float = np.nan, <- Added
    label: int = np.nan,
    score: float = np.nan,
    velocity: Tuple = (np.nan, np.nan),
    name: Optional[str] = None,
    token: Optional[str] = None
  ):
\end{lstlisting}

Further, we extend the class \texttt{NuScenesMetric} with \texttt{NuScenesMetricUQ} (see Listing~\ref{lst:nuscenes_metric_uq}) to include calibration and evaluation methods for classification and regression uncertainties.
It logs per-class uncertainty classification and regression metrics alongside the standard nuScenes mAP/NDS scores.

\begin{lstlisting}[language=Python, caption=Interface for NuScenes metric extension with calibration hooks and example usage., label=lst:nuscenes_metric_uq]
class NuScenesMetricUQ(NuScenesMetric):
  def __init__(
    self,
    *args,
    cls_calib_method='identity',
    reg_calib_method='identity',
    train_calibration=None,
    calib_dir=None,
    **kwargs
  ):

results = {...}  # List[NuScenesBoxUQ]

evaluator = NuScenesMetricUQ(...)

metrics = evaluator.nus_evaluate(
  results=results
)
\end{lstlisting}
The framework is directly compatible with \texttt{mmdetection3d}~\cite{mmdet3d2020} by using the redefined box structure and passing them to our evaluation routine.
We hope that this easy-to-use extension will facilitate future research on uncertainty calibration and evaluation for 3D object detection.

\PAR{Additional Metrics.}
We implement in our framework additional metrics for measuring the quality of uncertainties, that we did not present in the main paper due to space constraints.

\PAR{LaACE (Classification).}
The Location-Aware Adaptive Calibration Error (LaACE) mirrors LaECE but removes the discrete binning by directly comparing every detection's confidence with its location-aware quality term~\cite{kuzucu2024calibration}.
For each detection we compute $lq_j = 1 - \frac{\min(d_j, \tau)}{\tau}$ using the euclidean center-distance $d_j$ to the matched ground-truth (using the nuScenes TP threshold of $\tau=2\text{m}$) and set $lq_j=0$ for unmatched predictions.
Let $s_j$ denote the detector's confidence for detection $j$. It then reads
\begin{equation}
    \text{LaACE} = \frac{1}{n}\sum_{j=1}^{n}\left|s_j - lq_j\right|
\end{equation}
which yields a location-aware calibration error without the sampling variance of binning while still penalizing confident yet poorly localized detections. 
In our experiments, LaACE behaves similarly to LaECE but generally yields higher errors (see Table~\ref{tab:laace_vs_laece}), since LaECE is averaged over bins, whereas LaACE is a pointwise metric.
\input{tables/5_laece_vs_laace.tex}

\PAR{Uncertainty Realism (Regression).}
For the centroid regression uncertainty, we additionally evaluate the Uncertainty Realism Criterion~\cite{sicking2019} with a Mahalanobis distance-based statistical test that operates directly on the posterior predictive Gaussians.
Each matched detection $i$ is characterized by its mean vector $\boldsymbol{\mu}_i$ and covariance matrix $\boldsymbol{\Sigma}_i$, derived from the uncertainty mechanism, along with the ground-truth regression target $\mathbf{y}_{i,\text{gt}}$.
The squared Mahalanobis distance of the ground truth under the predictive Gaussian reads
\begin{equation}
    M_{\boldsymbol{\mu}_i,\boldsymbol{\Sigma}_i}^2(\mathbf{y}_{i,\text{gt}}) = (\mathbf{y}_{i,\text{gt}} - \boldsymbol{\mu}_i)^\top \boldsymbol{\Sigma}_i^{-1} (\mathbf{y}_{i,\text{gt}} - \boldsymbol{\mu}_i),
\end{equation}
and we collect the sample distances over the test set as $\mathcal{M}_{\text{gt}} = \{M_{\boldsymbol{\mu}_i,\boldsymbol{\Sigma}_i}^2(\mathbf{y}_{i,\text{gt}})\}_{i=1}^{N}$.

It reads $\chi^2(d)$ in distribution when the predictive mean and covariance are realistic, where $d$ is the dimensionality of the regression subspace (e.g., $d=3$ for centroid).

Following the Uncertainty Realism Criterion \cite{sicking2019}, the squared Mahalanobis distances for the entire test set $\mathcal{M}_{\text{gt}}$ should follow a $\chi^2(d)$-distribution.
We assess this hypothesis with the one-sample Kolmogorov--Smirnov test between $\mathcal{M}_{\text{gt}}$ and $\chi^2(d)$, but instead of using the highly sensitive p-value we record the KS statistic, which proved much more stable in practice.
Let $\hat{F}_{\mathcal{M}}$ denote the empirical CDF of $\mathcal{M}_{\text{gt}}$ and $F_{\chi^2(d)}$ the theoretical CDF of the $\chi^2(d)$ law.
Our reported score $\text{KS}_{xyz}$ corresponds to the maximum CDF deviation
\begin{equation}
    \text{KS}_{xyz} = D_{\text{KS}} = \sup_x \left| \hat{F}_{\mathcal{M}}(x) - F_{\chi^2(d)}(x) \right|,
\end{equation}
where lower values indicate that the predicted covariance matrices generate Mahalanobis distances closer to the desired $\chi^2(d)$ distribution.
We multiply $\text{KS}_{xyz}$ by 100 for better readability, resulting in values between 0 and 100, where 0 indicates perfect uncertainty realism and a value close to 100 indicates unrealistic uncertainties.
We report $\text{KS}_{xyz}$ alongside with MCA$_{xyz}$ in Table~\ref{tab:7_multicorrupt_reg_mean_pvalue} and since both metrics behave similarly across all methods, we conclude that MCA is a reliable proxy for regression Uncertainty Realism in our experiments.
A visual example of the empirical and theoretical CDFs for SECOND is provided in Figure~\ref{fig:second_mahalanobis}.
In our main experiments, we focus on MCA$_{xyz}$ as it is easier to interpret and more commonly used in the literature.
\input{figs/second_mahalanobis_figure.tex}
\input{tables/7_multicorrupt_reg_mean_pvalue.tex}

\subsection{Calibrator Parameter Count}\label{sec:calibrator_complexity}
\addcontentsline{aptoc}{subsection}{\protect\numberline{\thesubsection}Calibrator Parameter Count}
We summarize the number of learnable parameters for each classification and regression calibrator in Table~\ref{tab:calibrator_complexity} using SECOND at detection threshold 0.30 as an example.
For nuScenes with 10 object classes, global calibrators (\gbox) have ten times fewer parameters than class-wise calibrators (\cbox).
IR~\cite{ir} exhibits the highest complexity among all methods, as it employs a piecewise linear function for classification calibration with multiple segment parameters.
Note that the number of parameters for IR varies depending on the input distribution of the calibration set used for fitting, while other methods have a fixed parameter count.
\input{tables/3_calibrator_complexity.tex}

\subsection{Sample-based Approaches}
\addcontentsline{aptoc}{subsection}{\protect\numberline{\thesubsection}Sample-based Approaches}
Monte-Carlo Dropout (MCD)~\cite{mcd} and Deep Ensembles (DE)~\cite{de} performed poorly in our experiments, since the clustering of multiple stochastic forward passes into final detections is challenging in object-dense scenes.
We follow related work~\cite{pitropov2022} and cluster multiple 3D boxes across stochastic passes with DBSCAN before averaging the box parameters and deriving uncertainties from the sample statistics.
We tested a 3D IoU and the L1-Norm to compute the distance between boxes of the same class and provide ablations on DBSCAN's maximum distance threshold ($\epsilon$) for PETR in Table~\ref{tab:04_petr_samplebased}.
For MCD and DE, we use by default $n=6$ stochastic forward passes, and additionally provide results for $n=12$ using the best DBSCAN configuration.
For both, the accuracy (mAP, NDS) and uncertainty quality (D-ECE, MCA$_{xyz}$), the sample-based methods are underperforming compared to the single-forward pass approaches presented in the main paper.
Further, a higher number of stochastic passes does not improve the results, but increases the computational overhead.
Hence, we keep $n=6$ for all main experiments in the paper.
Compared to previous works~\cite{pitropov2022}, we apply MCD and DE to nuScenes, which is a more complex and object-dense dataset than KITTI, which contains only classes \texttt{car}, \texttt{pedestrian}, and \texttt{cyclist}.
Hence, both sample-based UQ baselines struggle to provide reliable uncertainty estimates for 3D object detection in object-dense scenes and produce mainly overconfident uncertainties, as shown in Figures~\ref{fig:qual_vis_de} and \ref{fig:qual_vis_mcd}.
This further underlines the need for dedicated uncertainty quantification methods in 3D object detection, as proposed in our work.
\input{tables/4_petr_samplebased}

\subsection{Train-time Calibration Approaches}
\addcontentsline{aptoc}{subsection}{\protect\numberline{\thesubsection}Train-time Calibration Approaches}

\PAR{CalDETR.}
Following the methodology of Cal-DETR~\cite{caldetr}, we adapt the training-time calibration framework for our DETR-based architecture.
We implement \textit{Uncertainty-Guided Logit Modulation} by estimating uncertainty via the variance of classification logits across the transformer decoder layers.
This variance modulates the final confidence scores during the forward pass, explicitly penalizing predictions that exhibit high disagreement across layers.
Furthermore, we incorporate \textit{Logit Mixing} as a regularizer, which interpolates the logits of positive queries with a batch-wise prototypical mean logit to enforce a smoother, well-calibrated embedding space.
For the Logit Mixing loss, we generate soft training targets where the ground-truth class receives a probability of $\alpha$, and the remaining probability mass $(1 - \alpha)$ is distributed among the other classes in the mix.
The loss is trained jointly with the standard 3D detection objectives (Classification Focal Loss, $L_1$).
We vary $\alpha$ as hyperparameter for PETR and SECOND and summarize the results in Table~\ref{tab:caldetr_alpha}.
Since $\alpha = 0.8$ gave us most stable results across both architectures, we use this value for all main experiments. 
We did not evaluate $\alpha < 0.5$ thoroughly, since the model focuses more on calibration and the NDS score drops significantly.
\input{tables/12_caldetr_alpha_table.tex}

\PAR{TCD.} We integrate the Train-time Calibration for Detection (TCD) loss~\cite{tcd} as an auxiliary differentiable objective within the transformer decoder layers to align predicted confidence with localization accuracy.
The loss formulation consists of two distinct components: a global classification alignment term and a local detection quality term.
For the classification component ($d_{cls}$), we compute the absolute difference between the mean predicted confidence (sigmoid probabilities) and the mean ground-truth accuracy (one-hot encoded labels) across the mini-batch.
This enforces that the average confidence reflects the empirical precision of the model for each class. For the detection component ($d_{det}$), we align the confidence of positive object queries with their structural fidelity.
We calculate the $L_1$ distance between the maximum class probability of a query and its corresponding Intersection-over-Union (IoU) with the ground truth box. 

The TCD loss is optimized jointly with the Focal Loss used for classification and the $L_1$ losses used for box regression.
The total TCD loss is defined as
\begin{equation}
    \mathcal{L}_{TCD} = 0.5 \cdot \left( \mathcal{L}_{cls}^{cal} + \mathcal{L}_{det}^{cal} \right).
\end{equation}
This auxiliary loss is applied at every decoder layer together with the standard detection losses.

\subsection{Density-Aware Calibration}
\addcontentsline{aptoc}{subsection}{\protect\numberline{\thesubsection}Density-Aware Calibration}
The gain parameter $\gamma$ controls how strongly the standardized density deviation modulates the per-query calibrators; larger values amplify the density-aware corrections (making the method more sensitive to distribution shift), while smaller values keep the adjustment closer to the base Post-hoc method behavior.
We therefore ablate $\gamma$ and summarize their effect on MCA$_{xyz}$ (see Table~\ref{tab:gamma_mca_petr_grid} and Table~\ref{tab:gamma_mca_second_grid}) and on D-ECE (see Table~\ref{tab:gamma_cls_petr_grid} and Table~\ref{tab:gamma_cls_second_grid}).
We tune $\gamma$ on the in-distribution validation set and fixed this value for the distribution-shift experiments.
A larger $\gamma$ allows the calibrator to react more aggressively to density drops, which is beneficial under strong shifts, but risks over-correcting on ID data.
For DA-TS (regression), we choose $\gamma=0.15$ for SECOND and $\gamma=0.3$ for PETR, as these values yield the best trade-off between ID and distribution-shift performance.
For classification, we chose $\gamma=0.2$ for both architectures and for both, DA-TS and DA-PS, as it provides a good trade-off between ID and distribution-shift performance.
The results indicate that a better tuning of $\gamma$ per architecture and calibrator could further improve the results under domain shift.
\input{tables/gamma_ablation_grid.tex}

\subsection{Distribution Shift Results}
\addcontentsline{aptoc}{subsection}{\protect\numberline{\thesubsection}Distribution Shift Results}
For SECOND, we present for each corruption the centroid regression quality metrics ($\text{MCA}_{xyz}$, $\text{KS}_{xyz}$) in Table~\ref{tab:multicorrupt_reg_second_xyz} and the remaining parameters ($\text{MCA}_{lwh}$, $\text{MCA}_{\phi}$) in Table~\ref{tab:multicorrupt_reg_second_lwhphi}.
For PETR, we present corresponding results in Tables~\ref{tab:multicorrupt_reg_petr_xyz} and~\ref{tab:multicorrupt_reg_petr_lwhphi}.
The reported values are averaged over three severity levels and detection thresholds from 0.05 to 0.60 in steps of 0.05.

\input{tables/2_second_multicorrupt_reg.tex}

\input{tables/3_second_multicorrupt_cls.tex}
\input{tables/2_petr_multicorrupt_reg.tex}
\input{tables/2_petr_multicorrupt_cls.tex}

\subsection{Qualitative Results}
\addcontentsline{aptoc}{subsection}{\protect\numberline{\thesubsection}Qualitative Results}
We provide qualitative examples, using PETR, of our density-aware calibration methods for an In-Distribution setting in Figures~\ref{qual_vis_uncal} and \ref{fig:qual_vis_calibrated}.
Additional qualitative results for DE~\cite{de} and MCD~\cite{mcd} are provided in Figures~\ref{fig:qual_vis_de} and \ref{fig:qual_vis_mcd}, respectively.
For distribution shift settings, we provide qualitative examples in Figures~\ref{fig:qual_vis_kluq_snow2}-\ref{fig:qual_vis_calibrated_brightness1}.
\input{figs/qualitative_vis.tex}

\FloatBarrier

\vfill\clearpage

%% file: tables/5_laece_vs_laace.tex
\begin{table*}[t]
\centering
\caption{
\textbf{In-Distribution - Classification Calibration}
LaACE and LaECE (in \%) are reported as mean over detections thresholds from $[0.05, 0.60]$ with 0.05 steps.
Class-wise calibration: \cbox; Global calibration: \gbox.
\textcolor{green!60!black}{+}/\textcolor{red!65!black}{-}: better/worse than the naïve post-hoc method.
}
\label{tab:laace_vs_laece}
\adjustbox{max width=15.0cm}{%
\begin{tabular}{w{c}{0.9cm}||c|w{c}{0.9cm}w{c}{1.1cm}|cc||w{c}{1.1cm}w{c}{1.2cm}|cc}
\toprule
\rowcolor{black!10!white} & & \multicolumn{4}{c||}{PETR} & \multicolumn{4}{c}{SECOND} \\
\midrule
Calib. & Calib. &  \multicolumn{2}{c|}{Calibration Error}     & \multicolumn{2}{c||}{Accuracy} & \multicolumn{2}{c|}{Calibration Error}  & \multicolumn{2}{c}{Accuracy} \\
Type   & Method & LaACE$\downarrow$ & LaECE$\downarrow$ &  mAP$\uparrow$ & NDS$\uparrow$ & LaACE$\downarrow$  &  LaECE$\downarrow$  &  mAP$\uparrow$  & NDS$\uparrow$ \\
\midrule
None & Uncal.             &  32.029   &     27.211 &                         38.25 &        45.05  &        27.136   &            17.378         & 54.53 & 63.57 \\
\midrule
\multirow{2}{*}{\shortstack{Sample\\-based}} 
 & MCD~\cite{mcd}         &  33.988            &      30.110          &      36.90 &        44.23  &        28.792   &                 19.523    & 51.80 & 62.30 \\
 & DE~\cite{de}           &  35.338            &      32.754          &      33.20 &        43.39  &        29.563   &                 19.635    & 50.83 & 62.36 \\
\midrule
\multirow{2}{*}{\shortstack{Train\\-time}}
 & CalDETR~\cite{caldetr} &  34.304            &      30.089          &      37.12 &         44.28  &       27.874   &                 19.906    & 53.57 & 63.17 \\
 & TCD~\cite{tcd}         &  28.837            &      24.533          &      38.27 &         45.05  &       28.696   &                 18.869    & 54.37 & 63.81 \\
\midrule
\multirow{6}{*}{\shortstack{Post\\-hoc}}
 & TS~\cite{ts} \gbox     &  31.156            &      26.115          &     38.26 &         45.04  &      23.175      &                 16.194   & 54.53 & 63.58 \\
 & TS~\cite{ts} \cbox     &  29.916            &      24.660          &     38.26 &         45.04  &      21.985      &                 13.968   & 54.53 & 63.58 \\
 & PS~\cite{ps} \gbox     &  30.770            &      25.686          &     38.26 &         45.04  &      22.932      &                 15.810   & 54.53 & 63.58 \\
 & PS~\cite{ps} \cbox     &  28.504            &      22.782          &     38.26 &         45.04  &      18.283      &                  9.076   & 54.53 & 63.58 \\
 & IR~\cite{ir} \gbox     &  30.534            &      25.753          &     38.26 &         45.04  &      22.591      &                 15.642   & 54.53 & 63.58 \\
 & IR~\cite{ir} \cbox     &  28.487            &      22.869          &     38.26 &         45.04  &      18.140      &                  9.078   & 54.53 & 63.58 \\
\midrule
 \multirow{6}{*}{\shortstack{Density\\-aware\\(Ours)}}
 & DA-TS \gbox           &  30.145\da{1}       & 24.831\da{4.92}      &    38.36 &        45.03  &       23.046\da{6.23}  &  15.185\da{6.23}      & 54.36 & 63.65 \\
 & DA-TS \cbox           &  29.546\da{1}       & 24.185\da{1.93}      &    38.36 &        45.03  &       21.796\da{6.23}  &  13.237\da{5.23}      & 54.36 & 63.65 \\
 & DA-PS \gbox           &  30.086\da{1}       & 24.960\da{2.83}      &    38.36 &        45.03  &       22.668\da{6.23}  &  15.043\da{4.85}      & 54.36 & 63.65 \\
 & \cb DA-PS \cbox           &  \cb\bf{28.104}\da{1}  & \cb\bf{22.393}\da{1.71} &  \cb38.36 &     \cb45.03  &  \cb\un{18.107}\da{6.23} &  \cb\bf{8.812}\da{2.91}  & \cb54.36 & \cb63.65 \\
 & DA-IR \gbox           &  29.226\da{1}       & 24.015\da{6.75}      &    38.36 &        45.03  &       18.872\da{6.23}  &   10.982\da{29.79}    & 54.36 & 63.65 \\
 & \cb DA-IR \cbox           &  \cb\un{28.160}\da{1}  & \cb\un{22.550}\da{1.39} &  \cb38.36 &     \cb45.03  &  \cb\bf{17.607}\da{6.23} &  \cb\un{8.877}\da{2.21}  & \cb54.36 & \cb63.65 \\
\bottomrule
\end{tabular}%
}%
\end{table*}

%% file: figs/second_mahalanobis_figure.tex
\begin{figure}[h]
\centering
\begin{subfigure}{0.32\columnwidth}
    \centering
    \includegraphics[width=\linewidth]{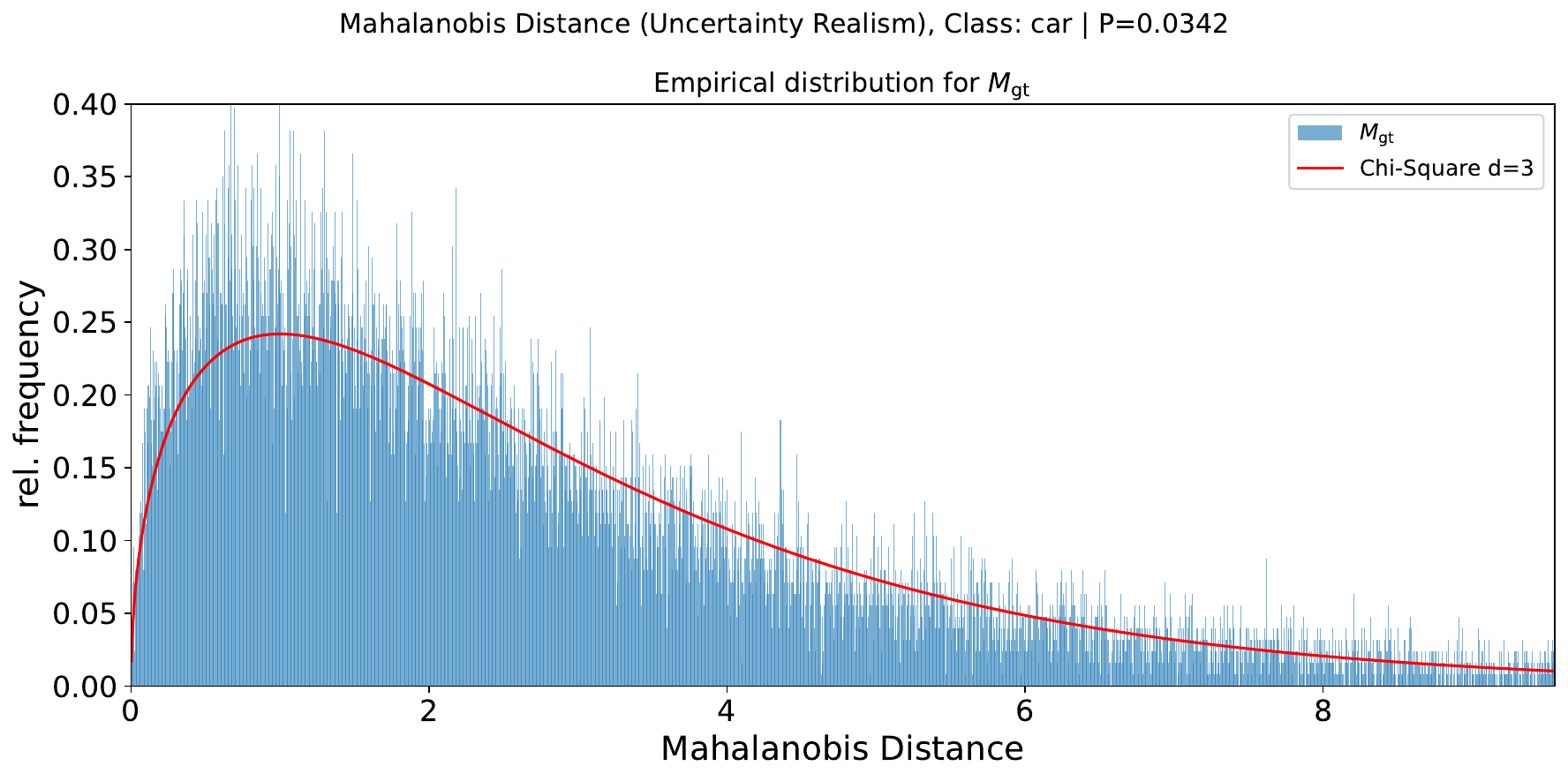}
    \caption{Well-Calibrated}
\end{subfigure}
\hfill
\begin{subfigure}{0.32\columnwidth}
    \centering
    \includegraphics[width=\linewidth]{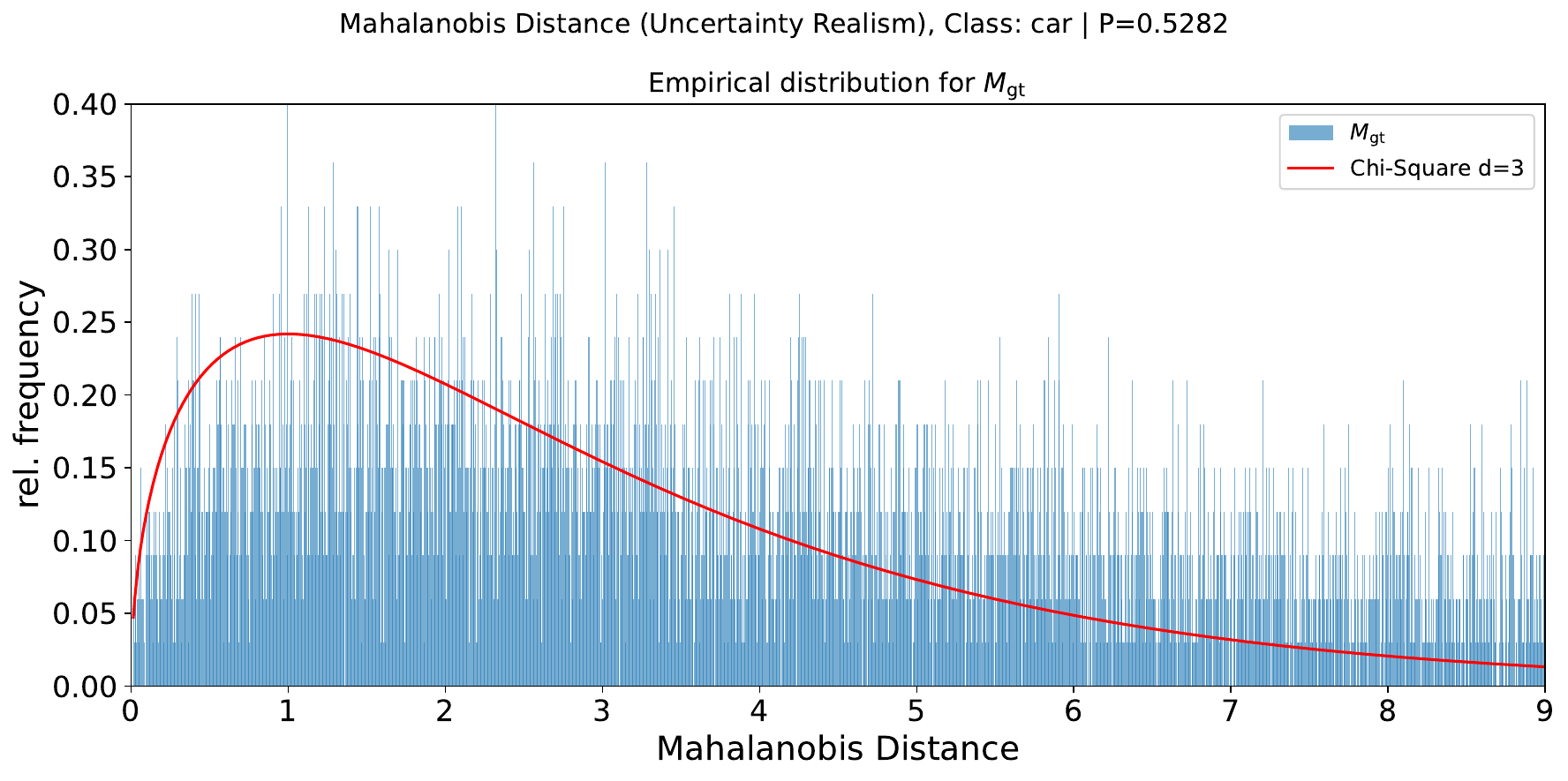}
    \caption{Underconfident}
\end{subfigure}
\hfill
\begin{subfigure}{0.32\columnwidth}
    \centering
    \includegraphics[width=\linewidth]{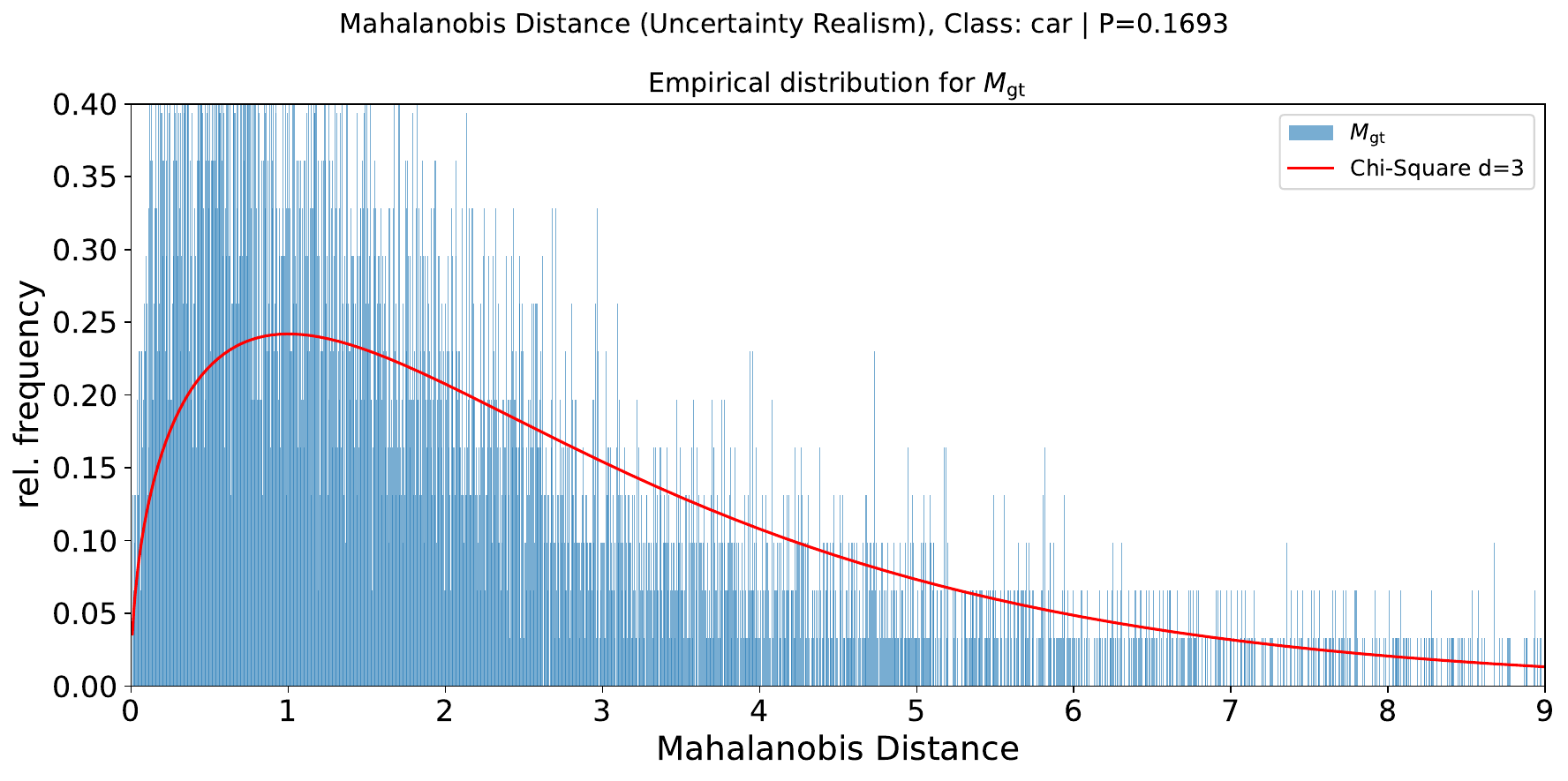}
    \caption{Overconfident}
\end{subfigure}
\caption{
\textbf{Uncertainty Realism Visualization.}
Empirical squared Mahalanobis Distance distributions (blue) and an ideal distribution (red) for the centroid covariance for a calibrated (a), overconfident (b) and underconfident case (c).}
\label{fig:second_mahalanobis}
\end{figure}

%% file: tables/7_multicorrupt_reg_mean_pvalue.tex
\begin{table*}[t]
\centering
\caption{
\textbf{In-Distribution - Regression Calibration}
Metrics are averaged over all corruptions, severity levels and detection thresholds from 0.05 to 0.60 with 0.05 increments.
Class-wise calibration: \cbox; Global calibration: \gbox.
}
\adjustbox{max width=15cm}{%
\begin{tabular}{c||c|cc|cc||cc|cc}
\toprule
\rowcolor{black!10!white} & & \multicolumn{4}{c||}{PETR} & \multicolumn{4}{c}{SECOND} \\
\midrule
Calib. &  Calib. & $\text{MCA}_{xyz}\!\downarrow$ & $\text{KS}_{xyz}\!\downarrow$ & mAP$\uparrow$ & NDS$\uparrow$ & $\text{MCA}_{xyz}\!\downarrow$ & $\text{KS}_{xyz}\!\downarrow$ & mAP$\uparrow$ & NDS$\uparrow$ \\
\midrule
\multirow{3}{*}{\shortstack{Sample\\-based}}
 & MCD~\cite{mcd}         & 35.878 & 92.003	& 36.90 & 44.23 & 31.620 & 87.506 & 51.80 & 62.30 \\
 & DE~\cite{de}           & 28.871 & 84.636 & 33.20 & 43.39 & 23.089 & 76.020 & 50.83 & 62.36 \\
 & CalDETR~\cite{caldetr} & 31.834 & 93.237	& 37.12 & 44.28 & 34.234 & 95.525 & 53.57 & 63.17 \\
\midrule
\multirow{2}{*}{\shortstack{Train\\-time}}
 & ES~\cite{es}           & 6.930  & 19.738 & 38.25 & 45.05 & 3.400  & 11.461 & 54.53 & 63.57 \\
 & KL~\cite{kld}          & 4.384  & 13.033 & 38.25 & 45.05 & 4.692  & 13.205 & 54.53 & 63.57 \\
\midrule
\multirow{4}{*}{\shortstack{Post\\-hoc}}
 & Depth \gbox            & 8.254  & 25.880 & 38.26 & 45.04 & 12.494 & 36.744 & 54.53 & 63.58 \\
 & Depth \cbox            & 2.407  & 10.881 & 38.26 & 45.04 & 4.749  & 19.102 & 54.53 & 63.58 \\
 & TS~\cite{trust} \gbox  & 4.028  & 11.977 & 38.26 & 45.04 & 3.443  & 12.711 & 54.53 & 63.58 \\
 & TS~\cite{trust} \cbox  & \un{1.692} & \un{7.135} & 38.26 & 45.04 & \un{1.533} & \un{7.281} & 54.53 & 63.58 \\
\midrule
\multirow{2}{*}{\shortstack{Density\\-aware}}
 & DA-TS \gbox            & 4.059\da{-1} & 12.087\da{-1} & 38.36 & 45.03 & 2.875\da{1} & 10.753\da{1} & 54.36 & 63.65 \\
 & DA-TS \cbox            & \cb\bf{1.538}\da{1} & \cb\bf{6.671}\da{1} & \cb38.36 & \cb45.03 & \cb\bf{1.518}\da{1} & \cb\bf{7.053}\da{1} & \cb54.36 & \cb63.65 \\
\bottomrule
\end{tabular}%
}%
\label{tab:7_multicorrupt_reg_mean_pvalue}
\end{table*}

%% file: tables/3_calibrator_complexity.tex
\begin{table}[h]
\centering
\caption{
\bf{Parameter Count of Post-hoc Calibration Methods.}
Parameter counts using SECOND for classification and regression calibration at detection threshold 0.30 and using ten object classes.}
\setlength{\tabcolsep}{6pt}
\begin{tabular}{llcc}
    \toprule
    Task & Method & \gbox & \cbox \\
    \midrule
    \multirow{6}{*}{\shortstack{Classi\\-fication}}
        & TS~\cite{ts} & 1 & 10 \\
        & PS~\cite{ps} & 2 & 20 \\
        & IR~\cite{ir} & 330 & 1364 \\
        & DA-TS & 2 & 20 \\
        & DA-PS & 4 & 40 \\
        & DA-IR & 377 & 1362 \\
    \midrule
    \multirow{3}{*}{\shortstack{Regress\\-ion}}
        & Depth           & 14 & 140 \\
        & TS~\cite{trust} & 7 & 70 \\
        & DA-TS           & 21 & 210 \\
    \bottomrule
\end{tabular}
\label{tab:calibrator_complexity}
\end{table}

%% file: tables/4_petr_samplebased.tex
\begin{table}[h]
\centering
\caption{
\bf{Sample-base Methods - PETR - In-Distribution.}
Accuracy (mAP and NDS) measured without thresholding.
D-ECE and MCA$_{xyz}$ measured at threshold $0.05$.
Bold indicates the configuration used for main experiments.
}
\resizebox{\columnwidth}{!}{%
\begin{tabular}{lcccccc}
\toprule
Method & Measure & $\epsilon$ & mAP$\uparrow$ & NDS$\uparrow$ & D-ECE$\downarrow$ & MCA$_{xyz}\!\downarrow$ \\
\midrule
\multirow{11}{*}{\shortstack{MCD \\ $n=6$ \\ \cite{mcd}}} & \multirow{5}{*}{\shortstack{3D IoU}} 
   & 0.50 & 35.53 & 43.94 & 9.67 & 36.30 \\
 & & 0.55 & 36.66 & 44.21 & 7.93 & 34.58 \\
 & & \bf{0.60} & \bf{36.57} & \bf{44.32} & \bf{9.70} & \bf{34.91} \\ %
 & & 0.65 & 36.65 & 44.29 & 9.76 & 34.44 \\
 & & 0.70 & 35.53 & 43.94 & 9.67 & 36.30 \\
 \cmidrule{3-7}
 & \multirow{5}{*}{\shortstack{L1}}
 & 0.60 & 29.86 & 44.19 & 9.29 & 36.88 \\
 & & 0.70 & 31.86 & 44.27 & 9.67 & 36.60 \\
 & & 0.75 & 33.07 & 44.31 & 8.45 & 35.72 \\
 & & 0.80 & 33.59 & 44.30 & 8.46 & 35.38 \\
 & & 0.90 & 34.43 & 44.16 & 8.30 & 34.73 \\
\midrule
 MCD $n=12$ \cite{mcd} & 3D IoU & 1.00 & 31.94 & 44.34 & 8.88 & 34.19 \\
\midrule
\midrule
\multirow{11}{*}{\shortstack{DE \\ $n=6$ \\ \cite{de}}} & \multirow{5}{*}{\shortstack{3D IoU}} 
   & 0.50 & 30.96 & 42.46 & 11.74 & 29.30 \\
 & & 0.55 & 32.45 & 42.92 & 11.67 & 27.80 \\
 & & 0.60 & 33.20 & 43.05 & 11.68 & 26.42 \\
 & & 0.65 & 33.33 & 42.82 & 11.46 & 25.053 \\
 & & 0.70 & 32.79 & 42.28 & 11.66 & 23.593 \\
 \cmidrule{3-7}
 & \multirow{5}{*}{\shortstack{L1}} 
   & 0.70 & 22.74 & 36.60 & 10.01 & 30.60 \\
 & & 0.75 & 23.89 & 40.40 & 10.35 & 30.11 \\
 & & 0.80 & 24.85 & 40.65 & 10.22 & 29.59 \\
 & & 0.90 & 25.51 & 42.47 & 10.98 & 28.78 \\ %
 & & \bf{1.00} & \bf{27.53} & \bf{43.39} & \bf{10.41} & \bf{27.48} \\
 & & 1.05 & 27.88 & 43.31 & 11.10 & 27.11 \\
 \midrule
 DE $n=12$ \cite{de} & L1 & 1.00 & 23.52 & 42.38 & 11.22 & 24.20 \\
\bottomrule
\end{tabular}
}
\label{tab:04_petr_samplebased}
\end{table}

%% file: tables/12_caldetr_alpha_table.tex
\begin{table}[h]
\centering
\caption{
\bf{CalDETR $\alpha$ Ablation - In-Distribution.}
D-ECE reported across detection thresholds from 0.05 to 0.60 with 0.05 increments.
Bold indicates the configuration used for main experiments.
}
\begin{tabular}{c|cc|cc}
  \toprule
  \multirow{2}{*}{$\alpha$} & \multicolumn{2}{c|}{SECOND} & \multicolumn{2}{c}{PETR} \\
    & D-ECE$\downarrow$ & NDS$\uparrow$ & D-ECE$\downarrow$ & NDS$\uparrow$ \\
  \midrule
        0.5 & 15.541 & 63.21 & 9.391 & 43.51 \\
   \bf{0.8} & 15.450 & 63.17 & 9.264 & 44.28 \\
        0.9 & 18.491 & 63.18 & 10.567 & 43.84 \\
  \bottomrule
\end{tabular}
\label{tab:caldetr_alpha}
\end{table}

%% file: tables/gamma_ablation_grid.tex
\begin{table*}[t]
    \centering
    \small
    \begin{tabular}{cc}
        \begin{minipage}{0.48\textwidth}
            \centering
            \captionof{table}{\textbf{Ablation on $\gamma$ - PETR - Regression.}}
            \label{tab:gamma_mca_petr_grid}
            \begin{tabular}{cc|cc}
                \toprule
                \multirow{2}{*}{Method} & \multirow{2}{*}{$\gamma$} & \multicolumn{2}{c}{MCA$_{xyz}\downarrow$} \\
                & & In-Distribution & Distribution-Shift \\
                \midrule
                \multirow{4}{*}{\shortstack{DA-TS \\ \cbox}}
                & 0.10          & 1.609 & 3.709 \\
                & 0.20          & 1.544 & 3.571 \\
                & \bf{0.30}     & 1.538 & 3.546 \\
                & 0.40          & 1.550 & 3.537 \\
                \midrule
                TS~\cite{trust}~\cbox & - & 1.692 & 3.780 \\
                \bottomrule
            \end{tabular}
        \end{minipage}
        &
        \begin{minipage}{0.48\textwidth}
            \centering
            \captionof{table}{\textbf{Ablation on $\gamma$ - SECOND - Regression.}}
            \label{tab:gamma_mca_second_grid}
            \begin{tabular}{cc|cc}
                \toprule
                \multirow{2}{*}{Method} & \multirow{2}{*}{$\gamma$} & \multicolumn{2}{c}{MCA$_{xyz}\downarrow$} \\
                & & In-Distribution & Distribution-Shift \\
                \midrule
                \multirow{4}{*}{\shortstack{DA-TS \\ \cbox}}
                &  0.05     & 1.526 & 4.584 \\
                &  0.10     & 1.538 & 4.241 \\ 
                &  \bf{0.15}& 1.518 & 4.037 \\
                &  0.20     & 1.589 & 4.031 \\
                \midrule
                TS~\cite{trust}~\cbox & - & 1.533 & 5.264 \\
                \bottomrule
            \end{tabular}
        \end{minipage}
        \\\\[10pt]
        \begin{minipage}{0.48\textwidth}
            \centering
            \captionof{table}{\textbf{Ablation on $\gamma$ - PETR - Classification.}}
            \label{tab:gamma_cls_petr_grid}
            \begin{tabular}{cc|cc}
                \toprule
                \multirow{2}{*}{Method} & \multirow{2}{*}{$\gamma$} & \multicolumn{2}{c}{D-ECE$\;\downarrow$} \\
                & & In-Distribution & Distribution-Shift \\
                \midrule
                \multirow{4}{*}{\shortstack{DA-TS \\ \cbox}}
                &  0.10     & 5.922 & 9.934 \\
                &  \bf{0.20}& 5.772 & 9.655 \\
                &  0.30     & 5.720 & 9.662 \\
                &  0.40     & 5.742 & 9.719 \\
                \midrule
                \multirow{4}{*}{\shortstack{DA-PS \\ \cbox}}
                &  0.10     & 4.202 & 7.978 \\
                &  \bf{0.20}& 3.899 & 7.770 \\
                &  0.30     & 3.968 & 7.764 \\
                &  0.40     & 4.115 & 7.801 \\
                \bottomrule
            \end{tabular}
        \end{minipage}
        &
        \begin{minipage}{0.48\textwidth}
            \centering
            \captionof{table}{\textbf{Ablation on $\gamma$ - SECOND - Classification.}}
            \label{tab:gamma_cls_second_grid}
            \begin{tabular}{cc|cc}
                \toprule
                \multirow{2}{*}{Method} & \multirow{2}{*}{$\gamma$} & \multicolumn{2}{c}{D-ECE$\;\downarrow$} \\
                & & In-Distribution & Distribution-Shift \\
                \midrule
                \multirow{4}{*}{\shortstack{DA-TS \\ \cbox}}
                &  0.10     & 7.793 & 13.461 \\
                &  \bf{0.20}& 7.351 & 13.018 \\
                &  0.30     & 7.189 & 12.901 \\
                &  0.40     & 7.035 & 12.746 \\
                \midrule
                \multirow{4}{*}{\shortstack{DA-PS \\ \cbox}}
                &  0.10     & 2.467 & 7.150 \\
                &  \bf{0.20}& 2.224 & 6.777 \\
                &  0.30     & 2.342 & 6.731 \\
                &  0.40     & 2.623 & 6.819 \\
                \bottomrule
            \end{tabular}
        \end{minipage}
    \end{tabular}
\end{table*}

%% file: tables/2_second_multicorrupt_reg.tex
\begin{table*}[t]
\centering
\caption{
\textbf{SECOND - Regression Calibration under Distribution Shift.}
Class-wise calibration (\cbox) and Global calibration (\gbox) reported for each corruption as calibration error $\text{MCA}_{xyz}$ and $\text{KS}_{xyz}$ (lower is better). Results are averaged over all detection thresholds and corruption severities.
Ours: DA-TS.
}
\setlength{\tabcolsep}{3pt}
\adjustbox{max width=\linewidth}{%
\begin{tabular}{lcccccccc|cccccccc}
\toprule
\multirow{2}{*}{Corruption} & \multicolumn{4}{c}{$\text{MCA}_{xyz}\downarrow$ (Calib: \cbox)} & \multicolumn{4}{c|}{$\text{MCA}_{xyz}\downarrow$ (Calib: \gbox)} & \multicolumn{4}{c}{$\text{KS}_{xyz}\downarrow$ (Calib: \cbox)} & \multicolumn{4}{c}{$\text{KS}_{xyz}\downarrow$ (Calib: \gbox)} \\
\cmidrule(lr){2-5}\cmidrule(lr){6-9}\cmidrule(lr){10-13}\cmidrule(lr){14-17}
            & KL\cite{kld} & Depth & TS~\cite{trust} & DA-TS &  KL\cite{kld} & Depth & TS~\cite{trust}  & DA-TS  &   KL\cite{kld}  & Depth & TS~\cite{trust} & DA-TS & KL\cite{kld}   & Depth & TS~\cite{trust} & DA-TS \\
\midrule
Snow        & 5.936 & 7.783  & 4.513 & \bf{2.237} &     5.936 & 14.997 & 4.966      & \bf{2.959}   &     15.426 & 26.635 &  11.788 &  \bf{6.881} & 15.426 & 44.587 & 14.204 & \bf{9.642} \\
Fog         & 6.389 & 7.905  & 5.164 & \bf{3.216} &     6.389 & 14.684 & 5.181 &     \bf{3.456}  &     16.297 & 27.358 & 13.799 &  \bf{9.018} & 16.297 & 43.745 & 14.850 & \bf{10.892} \\
Motionblur  & 4.824 & 7.875  & 3.528 & \bf{2.167} &     4.824 & 14.959 & 3.871      & \bf{2.744}   &     12.417 & 26.126 &  9.568 &  \bf{7.673} & 12.417 & 43.467 & 11.826 & \bf{9.732} \\
Beams Red.  & 6.820 & 12.638 & 5.952 & \bf{5.262} &     6.820 & 19.120 & 6.066      & \bf{5.470}   &     13.701 & 39.886 & 13.198 & \bf{15.652} & \bf{13.701} & 55.776 & 14.371 & 16.020 \\
Points Red. & 6.034 & 7.066  & 4.792 & \bf{2.388} &     6.034 & 14.446 & 5.045      & \bf{3.091}   &     15.753 & 25.526 &  12.936 &  \bf{7.266} & 15.753 & 43.371 & 13.720 & \bf{9.472} \\
Spatial Mis.& 8.172 & 16.117 & \bf{7.632} & 8.949 & \bf{8.172} & 22.092 & 8.987 & 9.908 & 25.491 & 45.918 & \bf{24.043} & 28.408 & \bf{25.491} & 59.510 & 27.643 & 31.072 \\
\midrule
Mean        & 6.363	 & 9.897  & 5.264 & \bf{4.037} &     6.363 & 16.716 & 5.686      & \bf{4.605}   &     16.514 & 31.908 &  14.222 &  \bf{12.483} & 16.514 & 48.410 & 16.103 & \bf{14.472} \\
\bottomrule
\end{tabular}
}%
\label{tab:multicorrupt_reg_second_xyz}
\end{table*}
\begin{table*}[t]
\centering
\caption{
\textbf{SECOND - Regression Calibration under Distribution Shift.}
Class-wise calibration (\cbox) and Global calibration (\gbox) reported for each corruption as calibration error $\text{MCA}_{lwh}$ and $\text{MCA}_{\phi}$ (lower is better). Results are averaged over all detection thresholds and corruption severities.
Ours: DA-TS.
}
\setlength{\tabcolsep}{3pt}
\adjustbox{max width=\linewidth}{%
\begin{tabular}{lcccccccc|cccccccc}
\toprule
\multirow{2}{*}{Corruption} & \multicolumn{4}{c}{$\text{MCA}_{lwh}\downarrow$ (Calib: \cbox)} & \multicolumn{4}{c|}{$\text{MCA}_{lwh}\downarrow$ (Calib: \gbox)} & \multicolumn{4}{c}{$\text{MCA}_{\phi}\downarrow$ (Calib: \cbox)} & \multicolumn{4}{c}{$\text{MCA}_{\phi}\downarrow$ (Calib: \gbox)} \\
\cmidrule(lr){2-5}\cmidrule(lr){6-9}\cmidrule(lr){10-13}\cmidrule(lr){14-17}
            & KL\cite{kld} & Depth & TS~\cite{trust} & DA-TS &  KL\cite{kld} & Depth & TS~\cite{trust}  & DA-TS  &   KL\cite{kld}  & Depth & TS~\cite{trust} & DA-TS & KL\cite{kld}   & Depth & TS~\cite{trust} & DA-TS \\
\midrule
Snow        & 5.722 & 6.881 & 5.049 & \bf{3.719} & 5.722 & 16.503 & 6.135 & \bf{5.161} & 17.530 & 8.127  & 9.364 & \bf{8.709} & 17.530 & 16.615 & 16.410 & \bf{10.923} \\
Fog         & 6.645 & 5.340 & 5.012 & \bf{3.516} & 6.645 & 15.506 & 7.685 & \bf{5.857} & 15.661 & 7.822  & 7.315 & \bf{7.194} & 15.661 & 16.738 & 14.727 & \bf{8.183} \\
Motionblur  & 5.574 & 6.358 & 4.354 & \bf{4.195} & 5.574 & 16.753 & 5.859 & \bf{6.092} & 14.453 & 7.996  & 7.578 & \bf{7.242} & 14.453 & 17.106 & 13.675 & \bf{7.935} \\
Beams Red.  & 7.969 & 9.896 & 6.848 & \bf{6.418} & 7.969 & 18.620 & 8.581 & \bf{8.397} & 15.924 & 10.773 & \bf{9.412} &      9.461 & 15.924 & 18.069 & 15.163 & \bf{10.344} \\
Points Red. & 6.378 & 3.675 & 4.850 & \bf{2.940} & 6.378 & 14.458 & 7.197 & \bf{5.308} & 15.944 & 9.076  & 7.973 & \bf{7.589} & 15.944 & 17.210 & 15.010 & \bf{9.132} \\
Spatial Mis.& 5.703 & 4.519 & 4.462 & \bf{2.859} & 5.703 & 15.346 & 6.695 & \bf{4.365} & 13.803 & 8.891  & 8.573 & \bf{8.437} & 13.803 & 15.399 & 12.107 & \bf{8.725} \\
\midrule
Mean        & 6.332 & 6.111 & 5.096 & \bf{3.941} & 6.332 & 16.198 & 7.025 & \bf{5.863} & 16.328 & 8.781  & 8.379 & \bf{8.105}  & 16.328 & 16.182 & 14.338 & \bf{9.207} \\
\bottomrule
\end{tabular}
}%
\label{tab:multicorrupt_reg_second_lwhphi}
\end{table*}

%% file: tables/3_second_multicorrupt_cls.tex
\begin{table*}[t]
\centering
\caption{
\textbf{SECOND - Classification Calibration under Distribution Shift.}
All results are reported for class-wise calibration (\cbox) and are averaged over all detection thresholds and all three corruption severities.
}
\setlength{\tabcolsep}{3pt}
\adjustbox{max width=14cm}{%
\begin{tabular}{lcccccc|cccccc}
\toprule
\multirow{3}{*}{Corruption} & \multicolumn{6}{c|}{D-ECE$\downarrow$ (Calib: \cbox)} & \multicolumn{6}{c}{LaECE$\downarrow$ (Calib: \cbox)} \\
            & \multicolumn{3}{c}{Baseline} & \multicolumn{3}{c|}{Density-aware (Ours)} & \multicolumn{3}{c}{Baseline} & \multicolumn{3}{c}{Density-aware (Ours)} \\
\cmidrule(lr){2-4}\cmidrule(lr){5-7}\cmidrule(lr){8-10}\cmidrule(lr){11-13}
             & TS\cite{ts} & PS\cite{ps} & IR\cite{ir} & DA-TS & DA-PS & DA-IR & TS\cite{ts} & PS\cite{ps} & IR\cite{ir} & DA-TS & DA-PS & DA-IR \\
\midrule
Snow         & 14.493 & 7.217  & 7.358  & \bf{7.316} & 7.403      & 8.993      & 16.448 & 8.987   & 9.153  & 13.278 & \bf{8.676} & 8.822 \\
Fog          & 13.054 & 6.492  & 6.364  & 12.581     & \bf{6.302} & 7.036      & 15.433 & 8.970   & 8.952  & 15.424 & 7.881      & \bf{7.441} \\
Motionblur   & 10.073 & 4.570  & 4.695  & 10.172     & \bf{4.408} & 4.887      & 14.256 & 8.725   & 8.770  & 14.547 & 8.110      & \bf{7.940} \\
Beams Red.   & 15.583 & 8.612  & 8.587  & 15.125     & 8.022      & \bf{6.785} & 18.223 & 11.734  & 11.655 & 13.287 & 7.451      & \bf{7.158} \\
Points Red.  & 10.606 & 4.513  & 4.406  & 10.374     & 4.635      & \bf{4.319} & 14.423 & 7.658   & 7.753  & 17.420 & 10.783     & \bf{8.514} \\
Spatial Mis. & 16.370 & 10.820 & 11.042 & 15.212     & 9.892      & 9.348      & 25.130 & 20.239  & 20.278 & 13.638 & \bf{7.243} & 7.288 \\
\midrule
Mean         & 13.363 & 7.037  & 7.076 & 13.018      & \bf{6.777} & 6.895      & 17.319 & 11.052  & 11.093 & 16.364 & 10.051     & \bf{9.273} \\
\bottomrule
\end{tabular}
}%
\label{tab:3_second_multicorrupt_cls}
\end{table*}

%% file: tables/2_petr_multicorrupt_reg.tex
\begin{table*}[h]
\centering
\caption{
\textbf{PETR - Regression Calibration under Distribution Shift.}
Class-wise calibration (\cbox) and Global calibration (\gbox) reported for each corruption as calibration error $\text{MCA}_{xyz}$ and $\text{KS}_{xyz}$ (lower is better). Results are averaged over all detection thresholds and corruption severities.
Ours: DA-TS.
}
\setlength{\tabcolsep}{3pt}
\adjustbox{max width=\linewidth}{%
\begin{tabular}{lcccccccc|cccccccc}
\toprule
\multirow{2}{*}{Corruption} & \multicolumn{4}{c}{$\text{MCA}_{xyz}\downarrow$ (Calib: \cbox)} & \multicolumn{4}{c|}{$\text{MCA}_{xyz}\downarrow$ (Calib: \gbox)} & \multicolumn{4}{c}{$\text{KS}_{xyz}\downarrow$ (Calib: \cbox)} & \multicolumn{4}{c}{$\text{KS}_{xyz}\downarrow$ (Calib: \gbox)} \\
\cmidrule(lr){2-5}\cmidrule(lr){6-9}\cmidrule(lr){10-13}\cmidrule(lr){14-17}
             & KL\cite{kld} & Depth & TS~\cite{trust} & DA-TS     & KL\cite{kld}    & Depth  & TS~\cite{trust} & DA-TS  & KL\cite{kld} & Depth & TS~\cite{trust}  & DA-TS    & KL\cite{kld} & Depth & TS~\cite{trust}   & DA-TS \\
\midrule
Snow         & 5.967 & 7.356 & 4.141 & \bf{3.599} & 5.967 & 13.063 & 5.811 & \bf{5.139} & 17.996 & 24.158 & 13.264 & \bf{12.018} & 17.996 & 39.370
 & 17.494 & \bf{15.294} \\
Fog          & 4.670 & 6.242 & 3.448 & \bf{3.149} & 4.670 &  11.775 & 4.539 & \bf{4.385} & 12.634 & 21.780 & 10.711 & \bf{9.734} & 12.634 & 36.086 & 12.117 & \bf{11.290} \\
Motionblur   & 5.062 & 8.349 & 3.618 & \bf{3.524} & 5.062 &  13.733 & 4.906 & \bf{4.444} & 13.538 & 26.861 & 11.115 & \bf{10.084} & 13.538 & 41.416 & 13.015 & \bf{12.773} \\
Brightness   & 5.572 & 9.220 & 4.781 & \bf{4.501} & 5.572 & 13.984 & 5.441 & \bf{5.390} & 14.565 & 28.557 & 14.342 & \bf{12.916} & 14.565 & 43.209 & 14.124 & \bf{14.079} \\
Darkness     & 4.236 & 5.665 & \bf{2.911} & 2.954 & 4.236 & 11.155 & \bf{4.151} & 4.319 & 11.628 & 20.615 & 9.748 & \bf{9.118} & 11.628 & 35.430 & 11.315 & \bf{11.607} \\
\midrule
Mean         & 5.101 & 7.367 & 3.780 & \bf{3.546} & 5.101 &  12.742 & 4.970 & \bf{4.735} & 14.072 & 24.394 & 11.836 & \bf{10.774} & 14.072 & 39.102 & 13.613 & \bf{13.009} \\
\bottomrule
\end{tabular}
}%
\label{tab:multicorrupt_reg_petr_xyz}
\end{table*}

\begin{table*}[h]
\centering
\caption{
\textbf{PETR - Regression Calibration under Distribution Shift.}
Class-wise calibration (\cbox) and Global calibration (\gbox) reported for each corruption as calibration error $\text{MCA}_{xyz}$ and $\text{KS}_{xyz}$ (lower is better). Results are averaged over all detection thresholds and corruption severities.
Ours: DA-TS.
}
\setlength{\tabcolsep}{3pt}
\adjustbox{max width=\linewidth}{%
\begin{tabular}{lcccccccc|cccccccc}
\toprule
\multirow{2}{*}{Corruption} & \multicolumn{4}{c}{$\text{MCA}_{lwh}\downarrow$ (Calib: \cbox)} & \multicolumn{4}{c|}{$\text{MCA}_{lwh}\downarrow$ (Calib: \gbox)} & \multicolumn{4}{c}{$\text{MCA}_{\phi}\downarrow$ (Calib: \cbox)} & \multicolumn{4}{c}{$\text{MCA}_{\phi}\downarrow$ (Calib: \gbox)} \\
\cmidrule(lr){2-5}\cmidrule(lr){6-9}\cmidrule(lr){10-13}\cmidrule(lr){14-17}
 & KL\cite{kld} & Depth & TS~\cite{trust} & DA-TS & KL\cite{kld} & Depth & TS~\cite{trust} & DA-TS & KL\cite{kld} & Depth & TS~\cite{trust} & DA-TS & KL\cite{kld} & Depth & TS~\cite{trust} & DA-TS \\
\midrule
Snow & 6.688 & 5.081 & 4.399 &  \bf{4.301} & 6.688 & 17.397 &  \bf{6.510} & 7.408 & 13.025 & 14.717 & 12.151 &  \bf{10.920} & 13.025 & 23.381 & 13.518 &  \bf{10.348} \\
Fog & 5.680 &  \bf{3.804} & 3.856 & 3.986 & 5.680 & 16.092 &  \bf{5.453} & 6.605 & 12.594 & 11.079 & 9.374 &  \bf{8.907} & 12.594 & 20.097 & 12.129 &  \bf{10.729} \\
Motionblur & 6.861 &  \bf{4.235} & 4.397 & 4.407 & 6.861 & 16.818 &  \bf{6.720} & 7.728 & 12.231 & 11.473 & 9.666 &  \bf{9.057} & 12.231 & 20.155 & 11.857 &  \bf{10.399} \\
Brightness & 6.305 & 6.189 & 4.829 &  \bf{4.698} & 6.305 & 17.036 &  \bf{6.100} & 6.395 & 11.640 & 12.988 & 10.819 &  \bf{9.764} & 11.640 & 21.218 & 11.820 &  \bf{9.932} \\
Darkness & 5.631 & 4.060 &  \bf{3.645} & 3.692 & 5.631 & 16.473 &  \bf{5.466} & 6.415 & 13.009 & 10.922 & 9.445 &  \bf{9.016} & 13.009 & 19.301 & 12.526 &  \bf{11.024} \\
\midrule
Mean         & 6.233 & 4.674 & 4.226 & \bf{4.217} & 6.233 & 16.763 & \bf{6.050} & 6.910 & 12.500 & 12.236  & 10.291 & \bf{9.533} & 12.500 & 20.830 & 12.370 & \bf{10.486} \\
\bottomrule
\end{tabular}
}%
\label{tab:multicorrupt_reg_petr_lwhphi}
\end{table*}

%% file: tables/2_petr_multicorrupt_cls.tex
\begin{table*}[t]
\centering
\caption{
\textbf{PETR - Classification Calibration under Distribution Shift.}
All results are reported for class-wise calibration (\cbox) and are averaged over all detection thresholds and all three corruption severities.
}
\setlength{\tabcolsep}{3pt}
\adjustbox{max width=14cm}{%
\begin{tabular}{lcccccc|cccccc}
\toprule
\multirow{3}{*}{Corruption} & \multicolumn{6}{c|}{D-ECE$\downarrow$ (Calib: \cbox)} & \multicolumn{6}{c}{LaECE$\downarrow$ (Calib: \cbox)} \\
            & \multicolumn{3}{c}{Baseline} & \multicolumn{3}{c|}{Density-aware (Ours)} & \multicolumn{3}{c}{Baseline} & \multicolumn{3}{c}{Density-aware (Ours)} \\
\cmidrule(lr){2-4}\cmidrule(lr){5-7}\cmidrule(lr){8-10}\cmidrule(lr){11-13}
            & TS\cite{ts} & PS\cite{ps} & IR\cite{ir} & DA-TS & DA-PS & DA-IR & TS\cite{ts} & PS\cite{ps} & IR\cite{ir} & DA-TS & DA-PS & DA-IR \\
\midrule
Snow        & 10.666 &  8.178 &  7.496      & 10.481 &  8.215 &  \bf{7.246} & 26.919 & 24.650 & 24.456 & 25.633 & 23.166 & \bf{22.201} \\
Fog         &  8.398 &  6.556 &  \bf{5.946} &  8.060 &  7.043 &  6.851      & 26.287 & 23.938 & 23.816 & 25.031 & 22.729 & \bf{21.820} \\
Motionblur  &  9.315 &  6.778 &  \bf{6.274} &  8.840 &  6.834 &  6.744      & 25.696 & 23.473 & 23.089 & 24.955 & 22.578 & \bf{21.629} \\
Brightness  & 11.788 &  9.551 &  9.137      & 11.757 &  9.784 &  \bf{8.709} & 28.557 & 26.028 & 25.676 & 27.906 & 24.983 & \bf{23.736} \\
Darkness    & 11.190 &  8.886 &  8.394      &  9.136 &  6.973 &  \bf{6.504} & 26.945 & 24.917 & 24.630 & 24.720 & 22.200 & \bf{21.675} \\
\midrule
Mean        & 10.271 &  7.990 &  7.449      &  9.655 &  7.770 &  \bf{7.211} & 26.881 & 24.602 & 24.333 & 25.649 & 23.131 & \bf{22.212} \\
\bottomrule
\end{tabular}
}%
\label{tab:2_petr_multicorrupt_cls}
\end{table*}

%% file: figs/qualitative_vis.tex
\begin{figure*}[htbp]
\centering
\includegraphics[trim=0cm 1cm 0cm 1cm, width=0.94\textwidth]{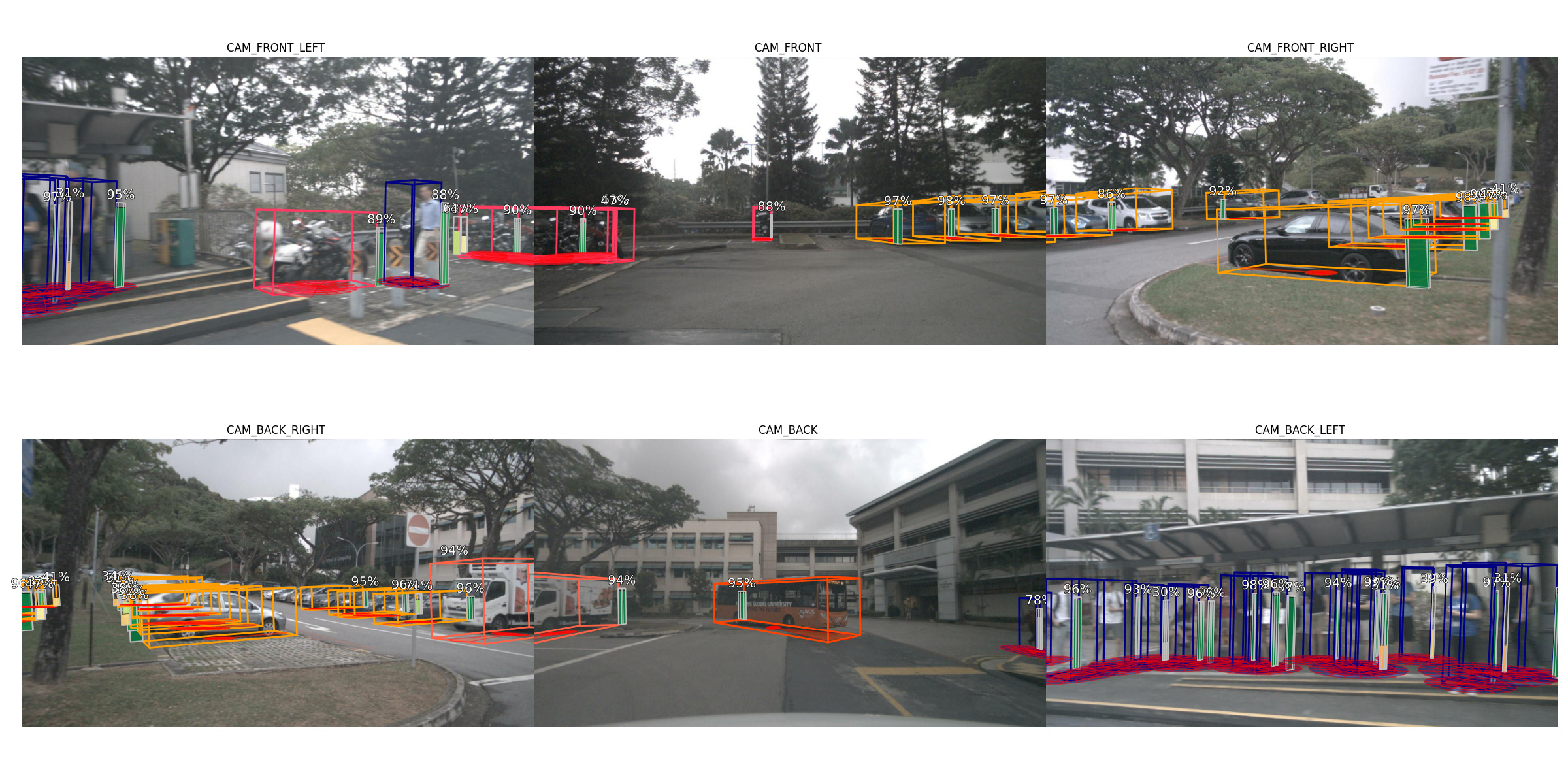}
\captionof{figure}{
\bf{In-Distribution}
PETR with KL~\cite{kld} - Uncalibrated.
}
\label{qual_vis_uncal}
\bigskip
\includegraphics[trim=0cm 1cm 0cm 1cm, width=0.94\textwidth]{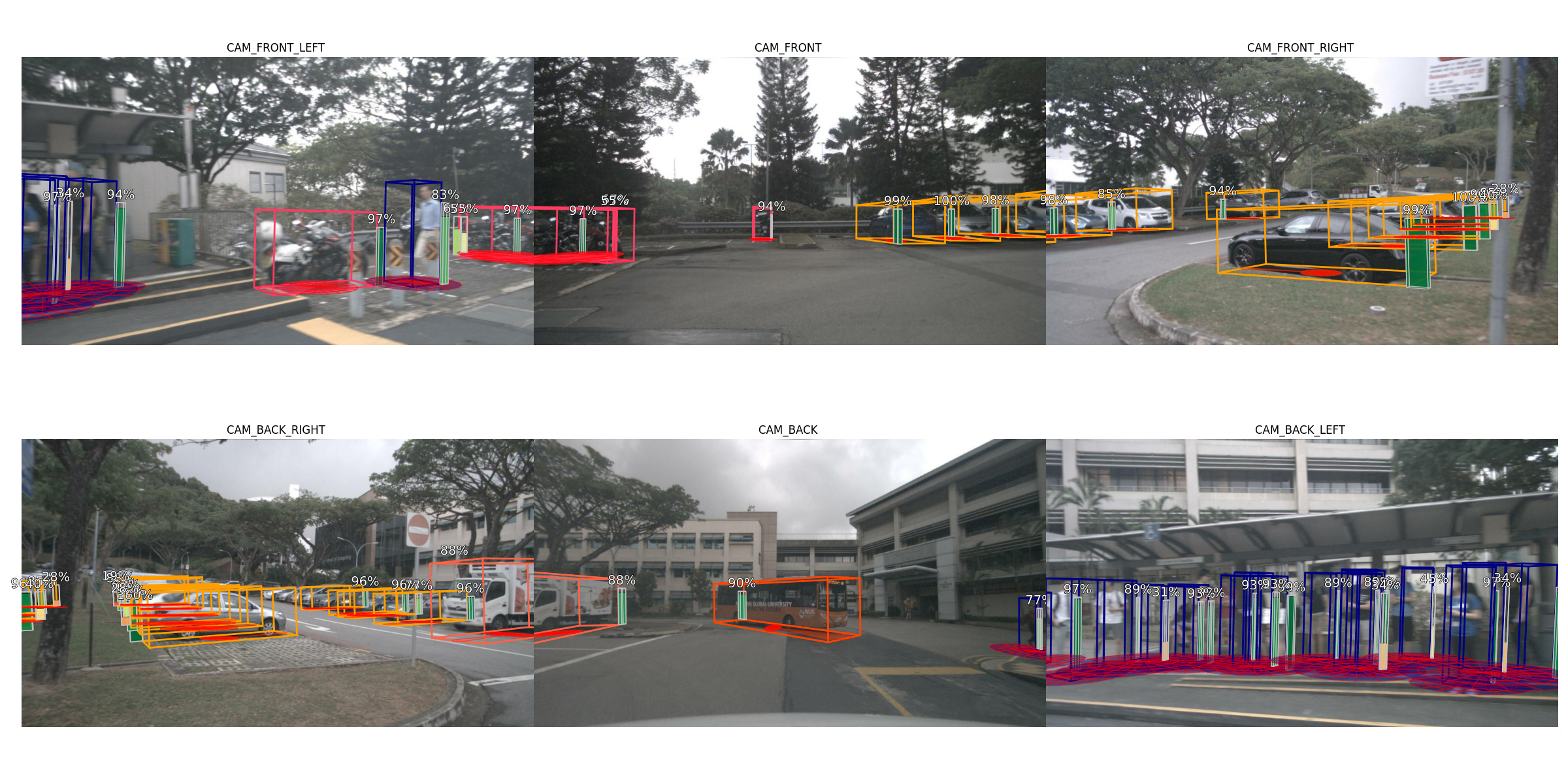}
\captionof{figure}{
\bf{In-Distribution}
PETR with KL~\cite{kld} calibrated with DA-TS for regression and DA-IR for classification.
}
\label{fig:qual_vis_calibrated}
\end{figure*}

\begin{figure*}[htbp]
\centering
\includegraphics[trim=0cm 1cm 0cm 1cm, width=0.94\textwidth]{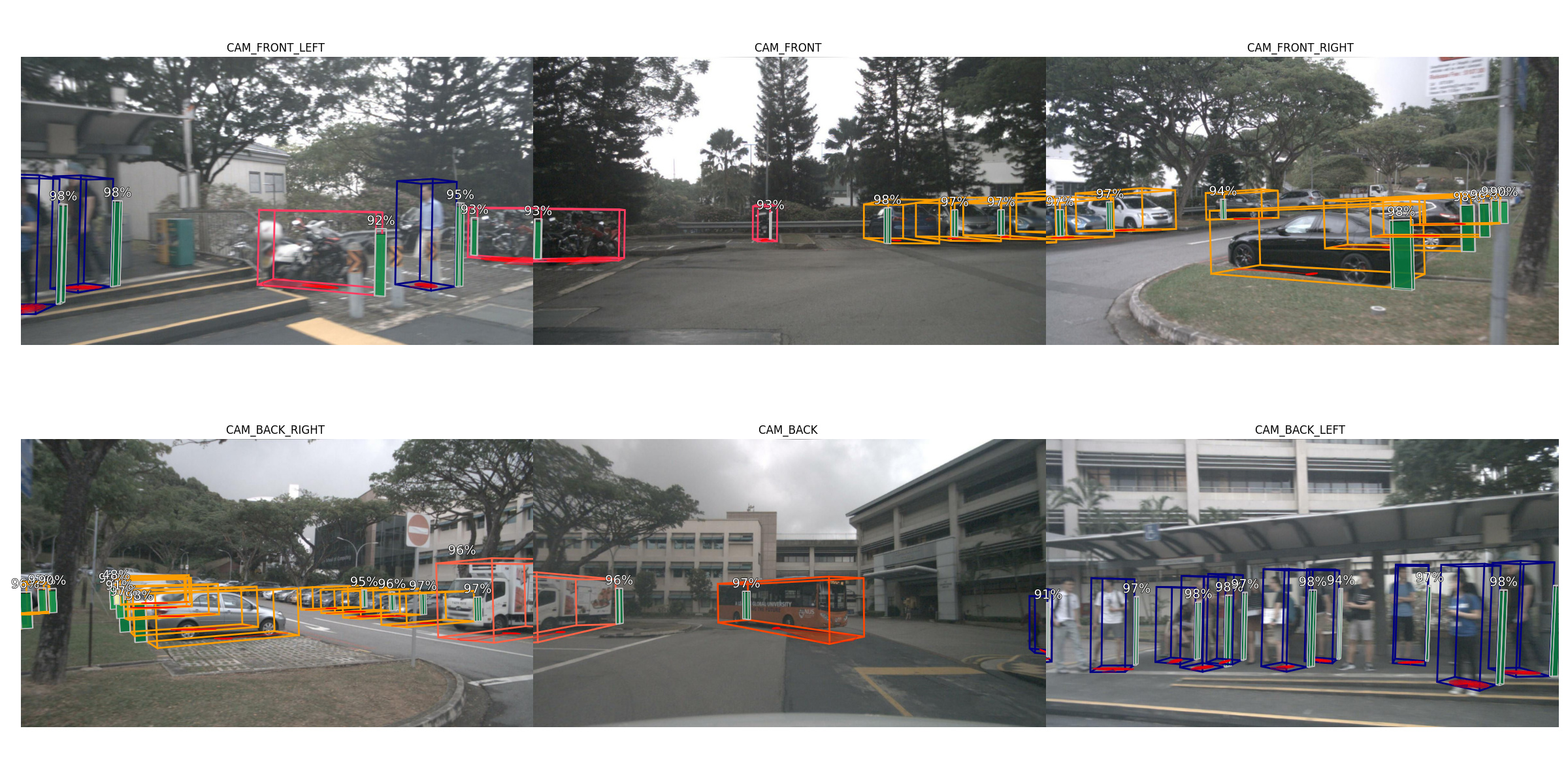}
\captionof{figure}{
\bf{In-Distribution}
PETR - DE~\cite{de}.
}
\label{fig:qual_vis_de}
\bigskip
\includegraphics[trim=0cm 1cm 0cm 1cm, width=0.94\textwidth]{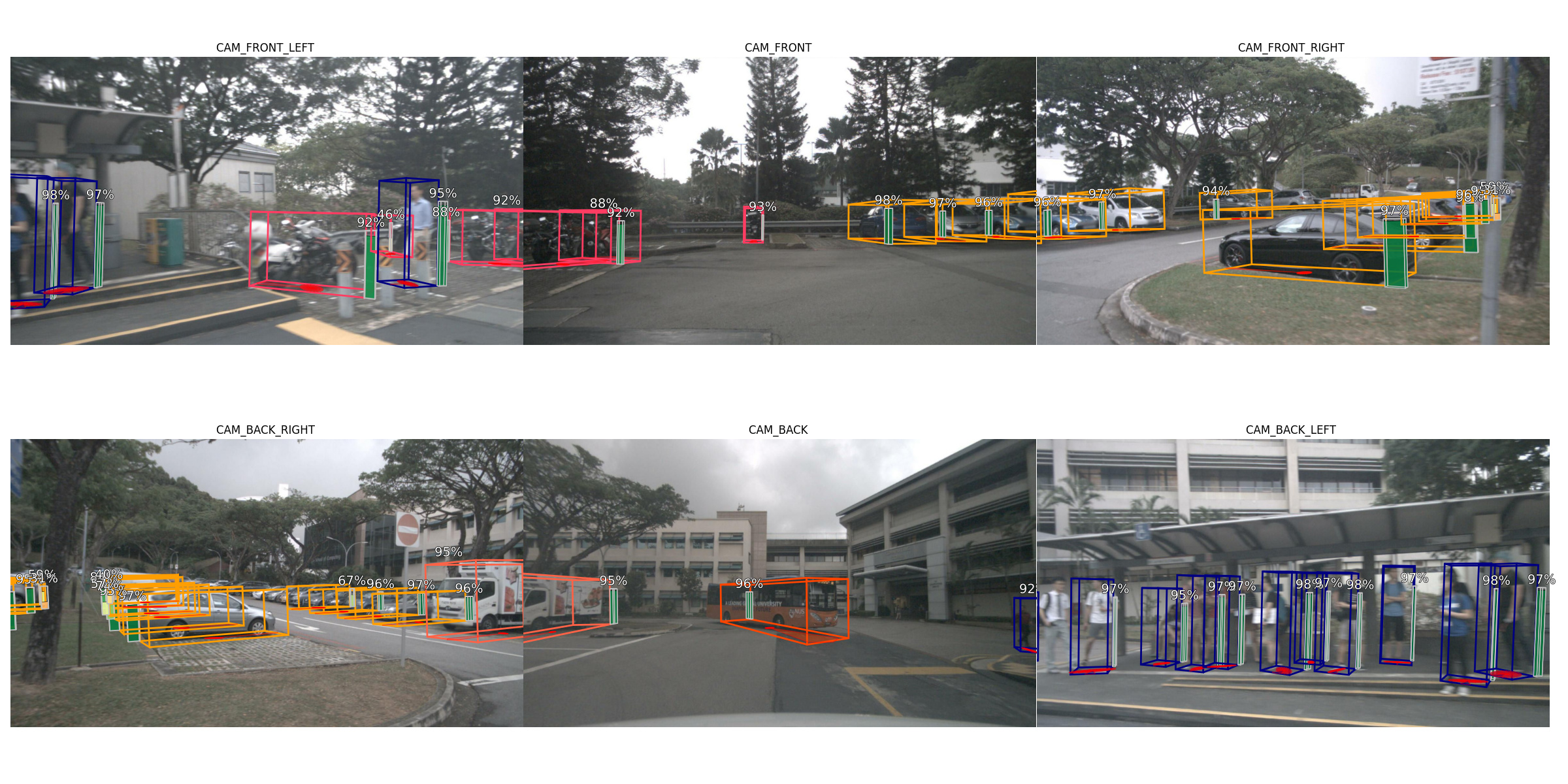}
\captionof{figure}{
\bf{In-Distribution}
PETR - MCD~\cite{mcd}.
}
\label{fig:qual_vis_mcd}
\end{figure*}

\begin{figure*}[htbp]
\centering
\includegraphics[trim=0cm 1cm 0cm 1cm, width=0.94\textwidth]{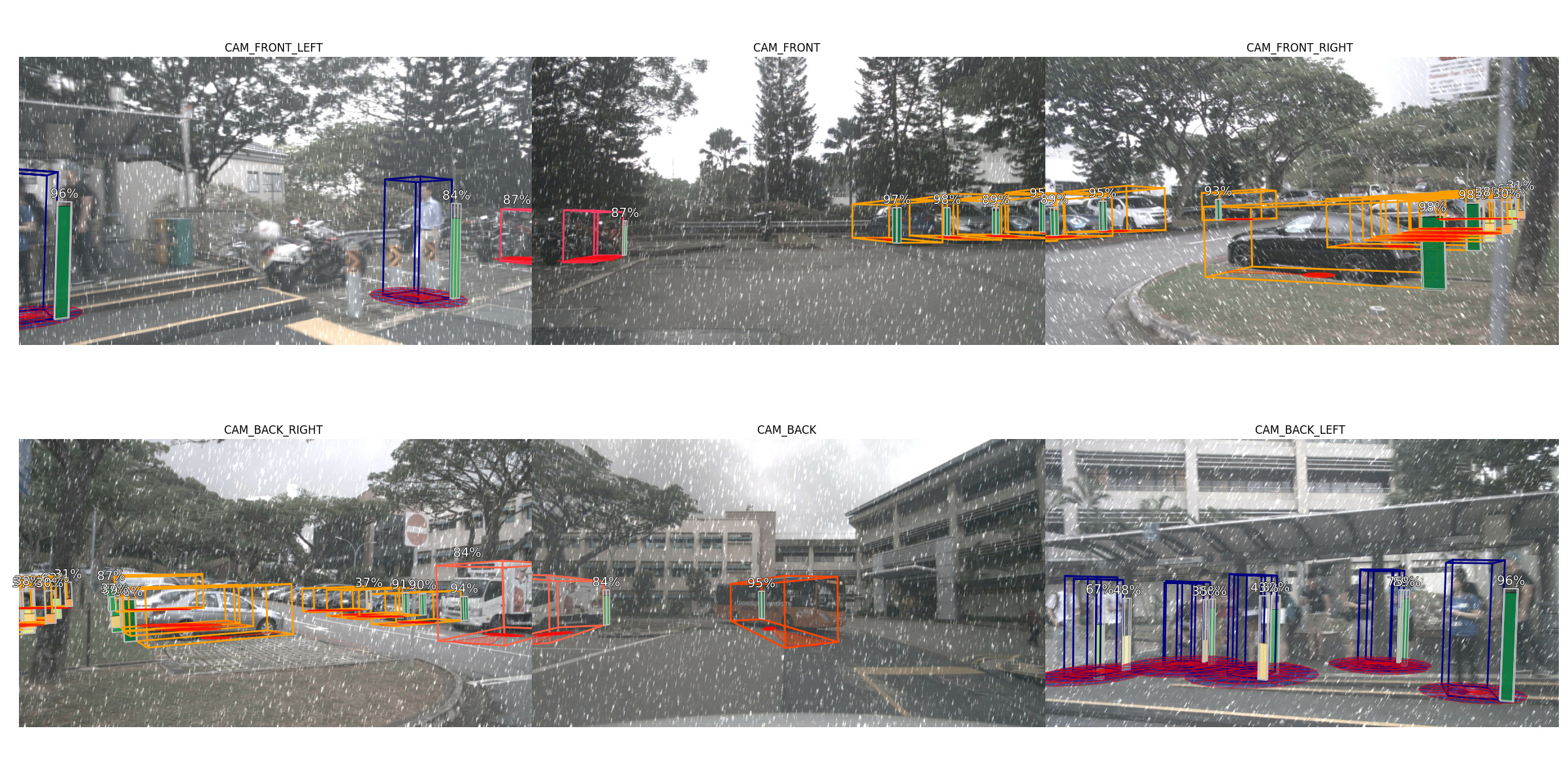}
\captionof{figure}{
\bf{Distribution Shift - Snow Level 2}
PETR with KL~\cite{kld} - Uncalibrated.
}
\label{fig:qual_vis_kluq_snow2}
\bigskip
\includegraphics[trim=0cm 1cm 0cm 1cm, width=0.94\textwidth]{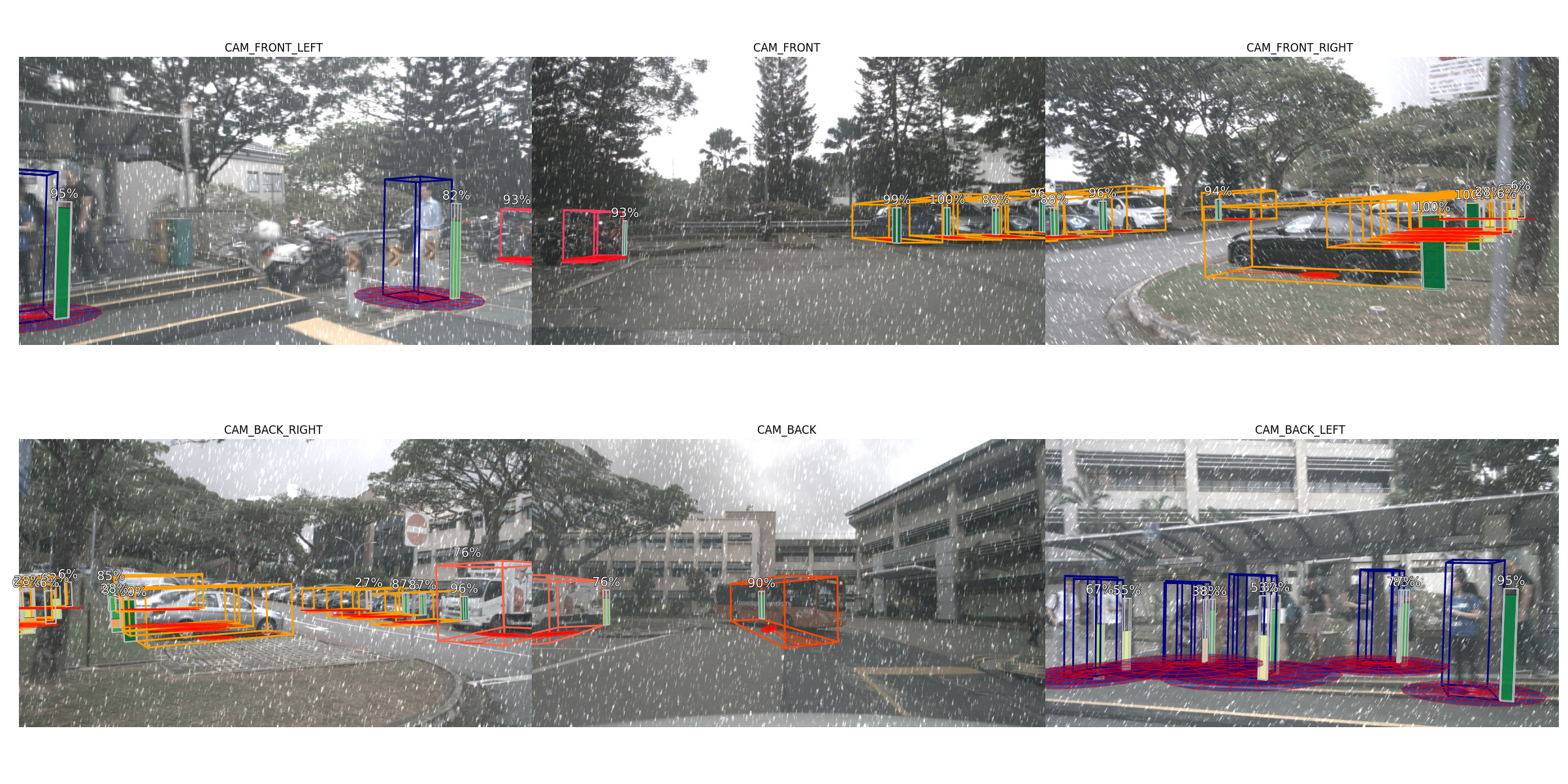}
\captionof{figure}{
\bf{Distribution Shift - Snow Level 2}
PETR with KL~\cite{kld} calibrated with DA-TS for regression and DA-IR for classification.
}
\label{fig:qual_vis_calibrated_snow2}
\end{figure*}

\begin{figure*}[htbp]
\centering
\includegraphics[trim=0cm 1cm 0cm 1cm, width=0.94\textwidth]{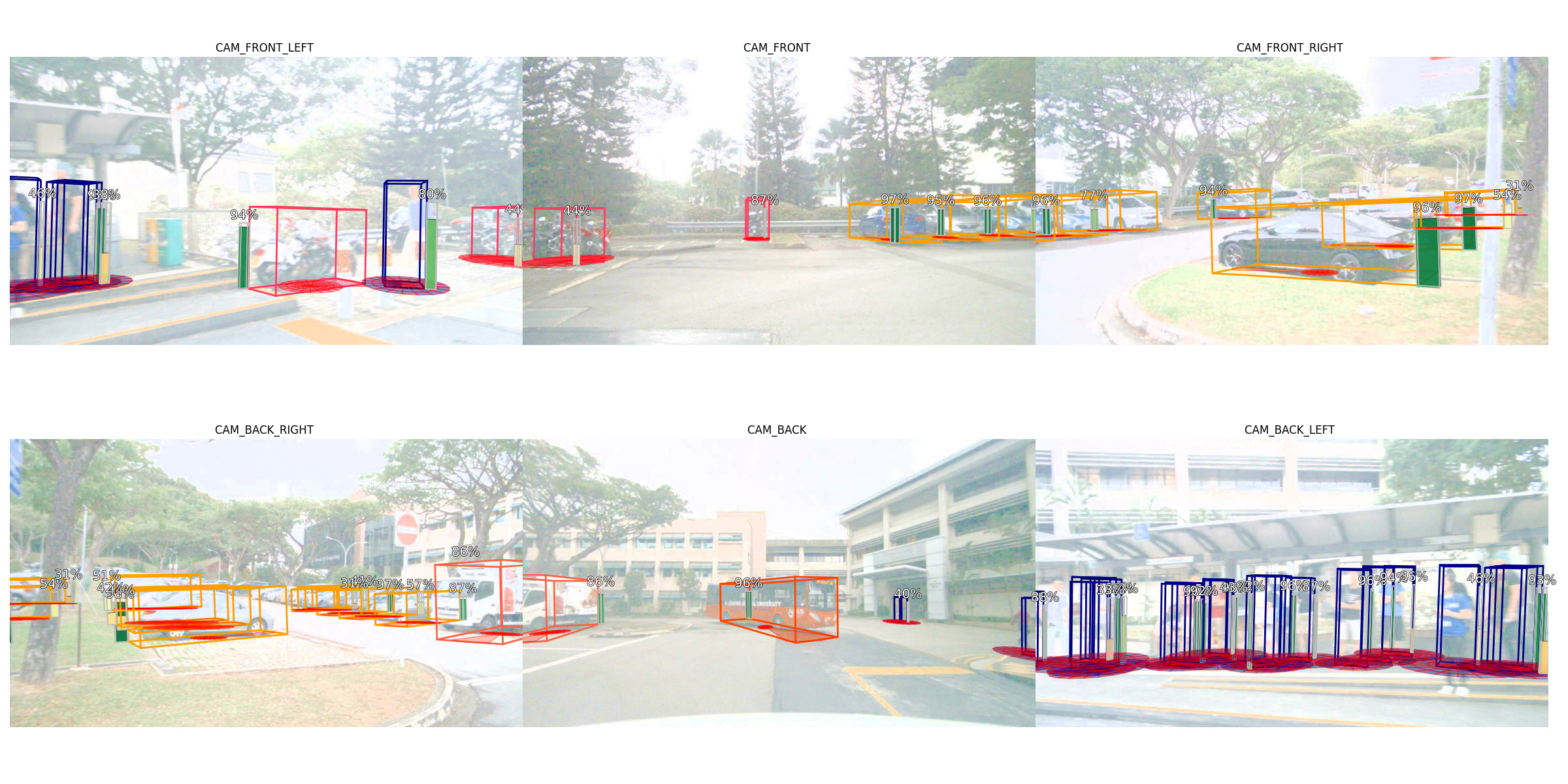}
\captionof{figure}{
\bf{Distribution Shift - Brightness Level 1}
PETR with KL~\cite{kld} - Uncalibrated.
}
\label{fig:qual_vis_kluq_brightness1}
\bigskip
\includegraphics[trim=0cm 1cm 0cm 1cm, width=0.94\textwidth]{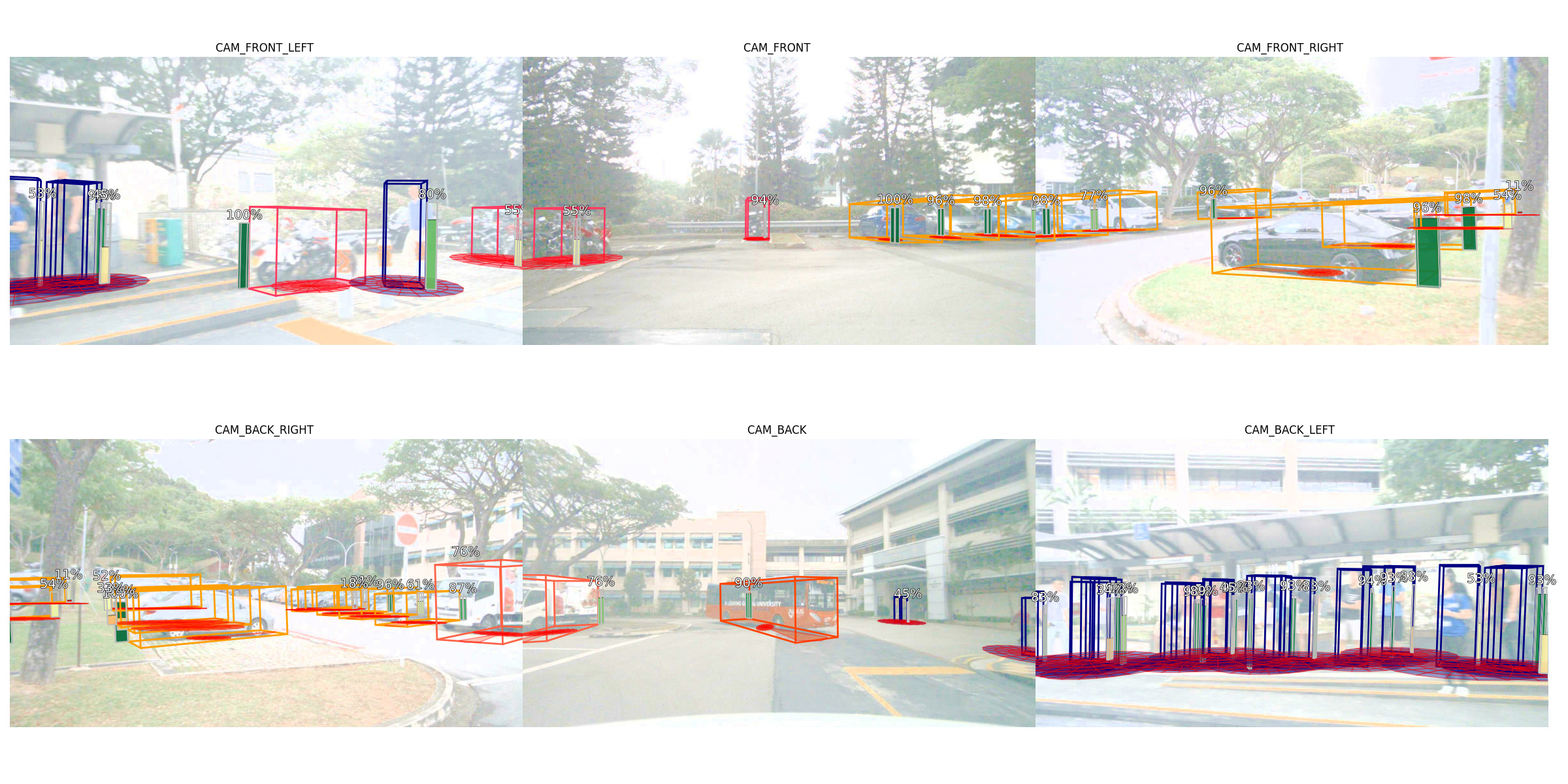}
\captionof{figure}{
\bf{Distribution Shift - Brightness Level 1}
PETR with KL~\cite{kld} calibrated with DA-TS for regression and DA-IR for classification.
}
\label{fig:qual_vis_calibrated_brightness1}
\end{figure*}